%% file: manuscript.tex
\documentclass{article}

\usepackage{microtype}
\usepackage{graphicx}
\usepackage{subfigure}
\usepackage{booktabs} 

\usepackage{hyperref}

\usepackage[accepted]{icml2025}

\usepackage{amsmath}
\usepackage{amssymb}
\usepackage{mathtools}
\usepackage{amsthm}

\usepackage[capitalize,noabbrev]{cleveref}

\theoremstyle{plain}

\theoremstyle{definition}

\theoremstyle{remark}

\usepackage[textsize=tiny]{todonotes}

\usepackage{hyperref}
\usepackage{url}
\usepackage{graphicx}
\usepackage{booktabs}
\usepackage{xcolor}
\usepackage{multirow}	  
\usepackage{tabularx}
\usepackage{caption}
\usepackage{subcaption}
\usepackage{wrapfig}
\usepackage{sidecap}
\usepackage{amssymb}
\usepackage{pifont}
\usepackage{adjustbox}
\usepackage{tikz}
\usepackage{stfloats}

\newcommand{\R}{\mathbb{R}}
\newcommand{\E}{\mathbb{E}}

\usepackage{acro}
\DeclareAcronym{MLP}{
  short = MLP,
  long = multilayer perceptron}
\DeclareAcronym{CNF}{
  short = CNF,
  long = continous normalizing flows} 
\DeclareAcronym{ODE}{
  short = ODE,
  long = ordinary differential equation}
\DeclareAcronym{NODE}{
  short = Neural ODE,
  long = neural ordinary differential equation}
\DeclareAcronym{PFM}{
  short = PFM,
  long = Pullback Flow Matching}
\DeclareAcronym{RFM}{
  short = RFM,
  long = Riemannian Flow Matching}
\DeclareAcronym{I-FABP}{
  short = I-FABP,
  long = intestinal fatty acid binding protein}
\DeclareAcronym{RMSE}{
  short = RMSE,
  long = root mean square error}
\DeclareAcronym{ARCH}{
  short = ARCH,
  long = ARCH}
\DeclareAcronym{VAE}{
  short = VAE,
  long = variational autoencoder}
\DeclareAcronym{RAE}{
  short = RAE,
  long = Riemannian Auto-Encoder}
\DeclareAcronym{GRAMPA}{
  short = GRAMPA,
  long = giant repository of AMP activities}
\DeclareAcronym{AK}{
  short = AK,
  long = Adenylate Kinase} 
\DeclareAcronym{NN}{
  short = NN,
  long = nearest neighbour}  
\DeclareAcronym{CFM}{
  short = CFM,
  long = Conditional Flow Matching}
\DeclareAcronym{FM}{
  short = FM,
  long = Flow Matching}  
\DeclareAcronym{OT}{
  short = OT,
  long = optimal transport}
\DeclareAcronym{MIC}{
  short = MIC,
  long = minimal inhibitory concentration}
\DeclareAcronym{KS}{
  short = KS,
  long = Kolmogorov-Smirnov}

\newcommand{\cmark}{\ding{51}}
\newcommand{\xmark}{\ding{55}}

\definecolor{kth-blue}{RGB/cmyk}{25,84,166/0.849,0.494,0,0.349}
\definecolor{tab_blue}{HTML}{1f77b4}
\definecolor{tab_orange}{HTML}{ff7f0e}
\definecolor{tab_green}{HTML}{2ca02c}
\definecolor{tab_purple}{HTML}{9467bd}
\definecolor{tab_red}{HTML}{d62728}

\icmltitlerunning{Pullback Flow Matching on Data Manifolds}

\begin{document}

\twocolumn[
\icmltitle{Pullback Flow Matching on Data Manifolds}



\icmlsetsymbol{equal}{*}

\begin{icmlauthorlist}
\icmlauthor{Friso de Kruiff}{kth,tudelft,uva}
\icmlauthor{Erik Bekkers}{uva}
\icmlauthor{Ozan Öktem}{kth}
\icmlauthor{Carola-Bibiane Schönlieb}{cam}
\icmlauthor{Willem Diepeveen}{ucla}
\end{icmlauthorlist}

\icmlaffiliation{kth}{Department of Mathematics, KTH Royal Institute of Technology}
\icmlaffiliation{uva}{AMLab, University of Amsterdam}
\icmlaffiliation{cam}{DAMTP, University of Cambridge}
\icmlaffiliation{ucla}{Department of Mathematics, University of California, Los Angeles}
\icmlaffiliation{tudelft}{Department of Applied Mathematics, Delft University of Technology}

\icmlcorrespondingauthor{Friso de Kruiff}{f.c.dekruiff@uva.nl}

\icmlkeywords{Isometric Learning, Pullback Geometry, Manifold Learning, Generative Modeling}

\vskip 0.3in
]



\printAffiliationsAndNotice{}  

\begin{abstract}
\input{sections/0_abstract}
\end{abstract}

\section{Introduction}
\label{sec:introduction}
\input{sections/1_introduction}

\section{Notation}
\label{sec:notation}
\input{sections/2_notation}

\section{Pullback Flow Matching}
\label{sec:pullback flow matching}
\input{sections/3_pullback_flow_matching}

\section{Learning Isometries}
\label{sec:learning isometries}
\input{sections/4_learning_isometries}

\section{Experiments}
\label{sec:experiments}
\input{sections/5_experiments}

\section{Conclusion}
\label{sec:conclusion}
\input{sections/6_conclusion}

\section*{Impact Statement}

This paper presents work whose goal is to advance the field of Machine Learning. There are many potential societal consequences of our work, none which we feel must be specifically highlighted here.

\section*{Acknowledgements}
FK acknowledges support from the ELLIS unit Amsterdam. EB acknowledges that this publication is part of the project SIGN with file number VI.Vidi.233.220 of the research programme Vidi which is (partly) financed by the Dutch Research Council (NWO) under the grant https://doi.org/10.61686/PKQGZ71565. OO acknowledges support from the Swedish Research Council 2020-03107. CBS acknowledges support from the Philip Leverhulme Prize, the Royal Society Wolfson Fellowship, the EPSRC advanced career fellowship EP/V029428/1, the EPSRC programme grant EP/V026259/1, and the EPSRC grants EP/S026045/1 and EP/T003553/1, EP/N014588/1, EP/T017961/1, the Wellcome Innovator Awards 215733/Z/19/Z and 221633/Z/20/Z, the EPSRC funded ProbAI hub EP/Y028783/1, the European Union Horizon 2020 research and innovation programme under the Marie Skodowska-Curie grant agreement REMODEL. This research was also supported by the NIHR Cambridge Biomedical Research Centre (NIHR203312). The views expressed are those of the author(s) and not necessarily those of the NIHR or the Department of Health and Social Care.

\bibliography{references}
\bibliographystyle{icml2025}

\newpage
\appendix
\onecolumn

\section{Background}
\label{sec:appendix background}
\input{sections/A.1_background}

\newpage

\section{Closed Form Manifold Mappings}
\label{sec:appendix closed form manifold mappings}
\input{sections/A.2_closed_form_manifold_mappings}

\section{Neural ODEs Parameterize Diffeomorphisms}
\label{sec:appendix neural odes parameterize diffeomorphisms}
\input{sections/A.3_proof_that_neural_odes_parameterize_a_diffeomorphisms}

\section{Manifold and Metric Selection}
\label{sec:appendix manifold and metric selection}
\input{sections/A.5_manifold_and_metric_selection}

\newpage

\section{Data Description}
\label{sec:appendix data description}
\input{sections/A.4_data_description}

\newpage

\section{Training Procedure}
\label{sec:appendix training procedure}
\input{sections/A.6_training_procedure}

\end{document}

%% file: sections/0_abstract.tex
We propose \ac{PFM}, a novel framework for generative modeling on data manifolds. Unlike existing methods that assume or learn restrictive closed-form manifold mappings for training \ac{RFM} models, \ac{PFM} leverages pullback geometry and isometric learning to preserve the underlying manifold's geometry while enabling efficient generation and precise interpolation in latent space. This approach not only facilitates closed-form mappings on the data manifold but also allows for designable latent spaces, using assumed metrics on both data and latent manifolds. By enhancing isometric learning through Neural ODEs and proposing a scalable training objective, we achieve a latent space more suitable for interpolation, leading to improved manifold learning and generative performance. We demonstrate \ac{PFM}'s effectiveness through applications in synthetic data, protein dynamics and protein sequence data, generating novel proteins with specific properties. This method shows strong potential for drug discovery and materials science, where generating novel samples with specific properties is of great interest.

%% file: sections/1_introduction.tex
Since the rise of machine learning in the scientific domain, researchers have focused on developing larger models trained on increasingly massive datasets, as in weather forecasting \citep{bodnar2024aurora} and protein structure prediction \citep{hayes2024simulating}. However, relying on such scaling laws is not feasible in many scientific fields where data is limited and precise modeling of physical phenomena is crucial. In such cases, incorporating prior knowledge about the geometry of the data as an inductive bias enables models to make accurate interpolations between data points, which is essential for reliable predictions and realistic representations of complex systems.

\begin{figure}[t]
    \centering
    \begin{minipage}{0.23\textwidth}
        \includegraphics[trim=0 0 0 50, clip, width=\textwidth]{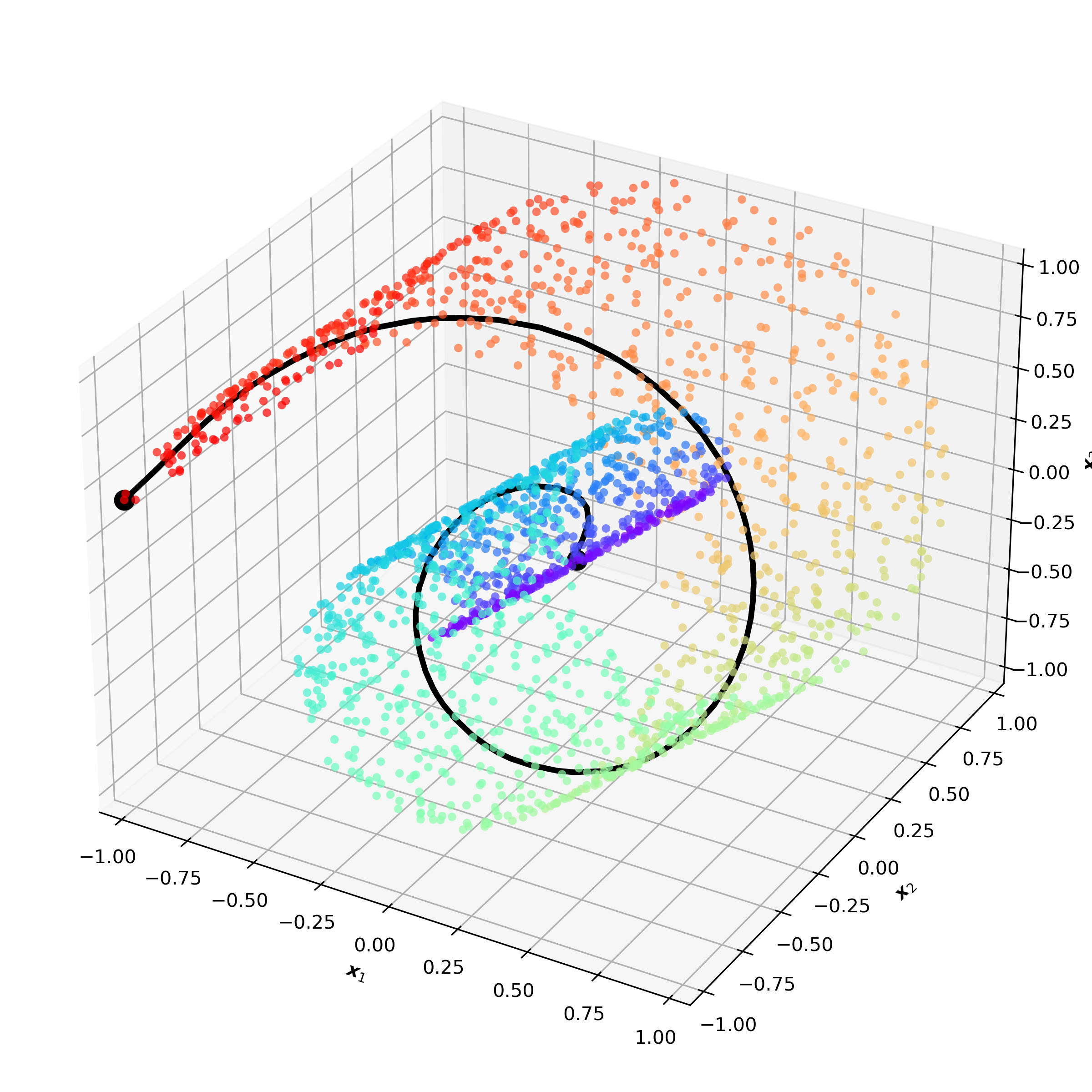}
        \textit{Data manifold} (\(\mathcal{D}\))
    \end{minipage}
    \hfill
    \begin{minipage}{0.23\textwidth}
        \includegraphics[trim=0 0 0 50, clip, width=\textwidth]{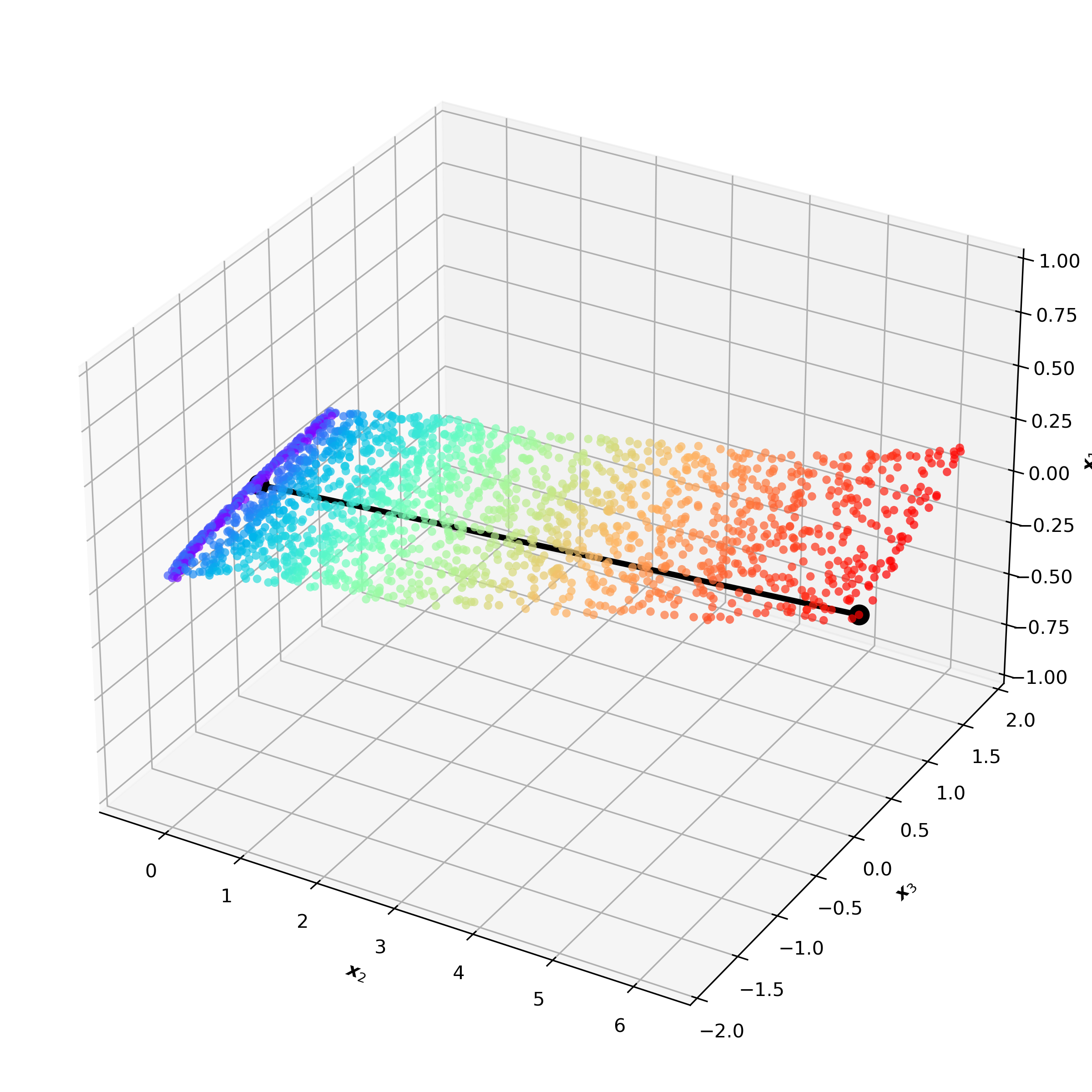}
        \textit{Latent manifold} (\(\mathcal{M}\))
    \end{minipage}
    \caption{Isometric learning for the rotated Swiss roll in 3D. The learned geodesic path (in black) on the data manifold $\mathcal{D} \subset \mathbb{R}^3$ correspond to the shortest paths on the latent manifold $\mathcal{M}=\R^3$.}
    \label{fig:swiss_roll_geodesic}
\end{figure}

Current methods, however, lack the mathematical foundations to accurately interpolate in latent space \citep{arvanitidis2017latent} and do not capture the underlying geometric structure of the data \citep{wessels2024grounding}. Our goal is to develop mappings that enable precise interpolation in latent space, leveraging geometry as an inductive bias to facilitate efficient and accurate generation on data manifolds, thereby advancing the ability to model complex physical phenomena with limited data. 

We consider modeling the data under the manifold hypothesis, which states that high-dimensional data lies on a lower dimensional manifold. This has been applied successfully to various scientific tasks \citep{vanderplas2009reducing,dsilva2016data,noe2017collective}. Modeling the data in its intrinsic dimension allows for efficient analysis \cite{diepeveen2024riemannian} and generation \cite{rombach2022high}. Furthermore, accurately capturing the geometry of the data manifold in the learning problem has shown to improve several down-stream tasks such as clustering \citep{ghojogh2022spectral}, classification \citep{kaya2019deep,hauberg2012geometric} and generation \citep{arvanitidis2020geometrically,sun2024geometry}. 

One way to achieve a latent manifold that supports interpolation is to have a structured Riemannian geometry, such as the one from \textit{pullback geometry}, which provides closed-form manifold mappings \citep{diepeveen2024riemannian}. This requires constructing an invertible and differentiable mapping---\textit{diffeomorphism}---between the data manifold and the latent manifold. Interpolation on manifolds is then performed through geodesics, shortest paths, and thus to achieve our goal we require geodesics on the data manifold to match geodesics on the latent manifold. This motivates our consideration of \textit{isometries}, that is, metric-preserving diffeomorphisms $\varphi$. These mappings preserve the distances of points on the data manifold on the latent manifold, and thereby ensure proper interpolation. 

\textbf{Related-Work}. In the literature, low-dimensional generation and generation on manifolds have typically been addressed as separate problems. Low-dimensional approaches, such as latent diffusion \citep{rombach2022high} or latent flow-matching \citep{dao2023flow}, often overlook the geometric structure of the data, leading to inaccuracies in tasks requiring a faithful representation of the underlying manifold. Conversely, manifold generation methods either assume geodesics on the data manifold for simulation-free training \citep{chen2024flow}—an approach inapplicable when closed-form mappings are unavailable—or attempt to learn a metric that forces the generative trajectories to have data support \citep{kapusniak2024metric}. 

Using a pullback framework presents challenges, such as task-specific learning problems that limit generality and prevent the learning of isometries across broader data manifolds \cite{cuzzolin2008learning,gruffaz2021learning,Lebanon2006}. Geometrically regularized latent space methods, like \cite{lee2022regularized} and \cite{duque2022geometry}, work in practice but lack solid mathematical grounding in isometries, particularly guaranteeing diffeomorphism in architectural design. \cite{diepeveen2024pulling} addresses isometry challenges with a more general mathematical framework, but its learning objective's expressivity and computational feasibility limit its application to high-dimensional real-world datasets.

Our approach bridges these gaps by modeling data on a lower-dimensional latent manifold with known geometry through diffeomorphisms parameterized and trained in a scalable and expressive way. By doing so we preserve the intrinsic properties of the data manifold and enable accurate and efficient generation through simulation-free training.

\textbf{Contributions.} We propose \ac{PFM}, a novel framework for latent manifold learning and generation through isometries. This method respects the geometry of the data manifold, even when closed-form manifold mappings are not available. Second, learning can be performed in the intrinsic dimension of the data manifold resulting in efficient and effective learning of the generative model with fewer parameters. We leverage pullback geometry to define a new metric on the entire ambient space, $\mathbb{R}^d$, by learning an isometry $\varphi$ that preserves the geometric structure of the data manifold $\mathcal{D}$ on the latent manifold $\mathcal{M}$. We use the corresponding metric of the assumed latent manifold $\mathcal{M}$ to perform \ac{RFM}. Our contributions are:
\begin{enumerate}
    \item We introduce \ac{PFM}, a novel framework that enables accurate and efficient data generation on manifolds. \Ac{PFM} leverages the pullback geometry to preserve the underlying geometric structure of the data manifold within the latent space, facilitating precise interpolation and generation.
    \item We improve the parameterization of diffeomorphisms, used to learn isometries, in both expressiveness and training efficiency through \acp{NODE}. 
    \item We introduce a scalable and stable isometric learning objective. This objective relies solely on a distance measure on the data manifold, simplifying the training process compared to \cite{diepeveen2024pulling} while maintaining geometric fidelity.
    \item We demonstrate our methods' effectiveness through experiments on synthetic data, high-dimensional molecular dynamics data, and experimental peptide sequences. Our framework utilizes \textit{designable latent spaces} to generate novel  proteins with specific properties closely matching reference samples. This directed generation showcases the significant applicability of isometric learning and \ac{PFM} in accurate physical modeling and interpolation, advancing generative modeling techniques in drug discovery and materials science. 
\end{enumerate}

%% file: sections/2_notation.tex
We give a brief summary of the notation used in the paper, and give a more extensive background on Riemannian and pullback geometry in \autoref{sec:appendix background}. 

A \textit{manifold} $\mathcal{M}$ is a topological space that locally resembles Euclidean space. A $d$-dimensional manifold $\mathcal{M}$ around a point $\boldsymbol{p} \in \mathcal{M}$ is described by a \textit{chart} $\psi: U \to \R^d$, where $U \subseteq \mathcal{M}$ is a neighborhood of $\boldsymbol{p}$. The chart provides a local coordinate system for the manifold. The \textit{tangent space} at a point $\boldsymbol{p} \in \mathcal{M}$, denoted $\mathcal{T}_{\boldsymbol{p}} \mathcal{M}$, is the vector space of all tangent vectors at that point.

A smooth manifold $\mathcal{M}$ equipped with a \textit{Riemannian metric} is called a \textit{Riemannian manifold} and is denoted by $(\mathcal{M}, (\cdot,\cdot)^{\mathcal{M}})$. The Riemannian metric $(\cdot, \cdot)^{\mathcal{M}}$ is a smoothly varying inner product defined on the tangent spaces $\mathcal{T}_{\boldsymbol{p}} \mathcal{M}$ for all points $\boldsymbol{p} \in \mathcal{M}$, and it defines lengths and angles on the manifold. A \textit{geodesic}, $\gamma_{\boldsymbol{p},\boldsymbol{q}}(t)$ is the shortest path between two points $\boldsymbol{p},\boldsymbol{q} \in \mathcal{M}$, generalizing the notion of a straight line in Euclidean space. 

The \textit{exponential map} $\exp_{\boldsymbol{p}}: \mathcal{T}_{\boldsymbol{p}}\mathcal{M} \to \mathcal{M}$ maps a tangent vector $\Xi_{\boldsymbol{p}}$ to a point on the manifold by following the geodesic in the direction of $\Xi_{\boldsymbol{p}}$ starting from $\boldsymbol{p}$. The inverse of the exponential map is the \textit{logarithmic map}, denoted by $\log_{\boldsymbol{p}}: \mathcal{M} \to \mathcal{T}_{\boldsymbol{p}}\mathcal{M}$, which returns the tangent vector corresponding to a given point on the manifold.

In this work, we consider a $d$-dimensional Riemannian manifold $\big(\mathcal{M}, (\cdot,\cdot)^\mathcal{M}\big)$, and a smooth diffeomorphism $\varphi: \R^d \rightarrow \mathcal{M}$, such that $\varphi\big(\R^d\big) \subseteq \mathcal{M}$ is geodesically convex, meaning that any pair of points within this subset are connected by a unique geodesic. 

This mapping allows us to pullback the geometric structure of $\mathcal{M}$ to $\R^d$ by defining the \textit{pullback metric} on $\R^d$. Specifically, for tangent vectors $\Xi_{\boldsymbol{p}}, \Phi_{\boldsymbol{p}} \in \mathcal{T}_{\boldsymbol{p}}\R^d$, the pullback metric is defined as
\begin{equation}
    (\Xi_{\boldsymbol{p}}, \Phi_{\boldsymbol{p}})^\varphi := \big(\varphi_\ast[\Xi_{\boldsymbol{p}}], \varphi_\ast[\Phi_{\boldsymbol{p}}]\big)_{\varphi(\boldsymbol{p})}^\mathcal{M},
\end{equation}
where $\varphi_\ast$ is the pushforward of tangent vectors under $\varphi$. Through this construction, various geometric objects in $\mathcal{M}$, such as distances and geodesics, can be expressed in terms of their counterparts in $\R^d$ with respect to the pullback metric. 

The distance function $d^\varphi_{\R^d}: \R^d \times \R^d \rightarrow \R$ on $\R^d$ with the pullback metric is given by,
\begin{equation}
    d^\varphi_{\R^d}(\boldsymbol{x}_i, \boldsymbol{x}_j) = d_\mathcal{M}\big(\varphi(\boldsymbol{x}_i), \varphi(\boldsymbol{x}_j)\big),
\end{equation}
where $d_\mathcal{M}$ denotes the Riemannian distance on $\mathcal{M}$. The length-minimizing geodesic connecting $\boldsymbol{x}_i$ and $\boldsymbol{x}_j$ in $\R^d$ with respect to the pullback metric $\gamma^\varphi_{\boldsymbol{x}_i, \boldsymbol{x}_j}: [0,1] \rightarrow \R^d$ is given by,
\begin{equation}
    \gamma^\varphi_{\boldsymbol{x}_i, \boldsymbol{x}_j}(t) = \varphi^{-1}\big(\gamma^\mathcal{M}_{\varphi(\boldsymbol{x}_i), \varphi(\boldsymbol{x}_j)}(t)\big), \label{eq:pullback geodesic equation}
\end{equation}
here $\gamma^\mathcal{M}$ denotes the geodesic in $\mathcal{M}$ connecting $\varphi(\boldsymbol{x}_i)$ and $\varphi(\boldsymbol{x}_j)$. This enables computation of geodesics and distances in $\R^d$ using the geometry of $\mathcal{M}$, as stated in Prop. 2.1 of \cite{diepeveen2024pulling}.

In this paper we will assume the standard Euclidean metric $(\cdot,\cdot)_2$ and a Euclidean latent manifold $\mathcal{M}=\R^d$. Hence, the pullback metric will be defined as
\begin{equation}
    (\Xi_{\boldsymbol{p}}, \Phi_{\boldsymbol{p}})^\varphi := \big(\varphi_\ast[\Xi_{\boldsymbol{p}}], \varphi_\ast[\Phi_{\boldsymbol{p}}]\big)_{\varphi(\boldsymbol{p})}^{\R^d}.
\end{equation}
We will calculate distances on the latent manifold $\mathcal{M}=\R^d$ through,
\begin{equation}
    d^\varphi_{\R^d}(\boldsymbol{x}_i, \boldsymbol{x}_j) = \|\varphi(\boldsymbol{x}_i)- \varphi(\boldsymbol{x}_j)\|_2,
\end{equation}
and the geodesic calculation will boil down to
\begin{equation}
     \gamma^\varphi_{\boldsymbol{x}_i, \boldsymbol{x}_j}(t) = \varphi^{-1}\big(\varphi(\boldsymbol{x}_i)(1-t) +t\varphi(\boldsymbol{x}_j)\big).
\end{equation}
An example of a pullback geodesic $\gamma^\varphi_{\boldsymbol{x}_i, \boldsymbol{x}_j}(t)$ on the data manifold based on a geodesic on a latent Euclidean manifold can be viewed in \autoref{fig:swiss_roll_geodesic}. 

\begin{figure*}[t!]
    \centering
    \begin{tikzpicture}
        \node[anchor=center] at (0, 0) {
            \includegraphics[trim=0 15 50 15, clip, width=0.75\linewidth]{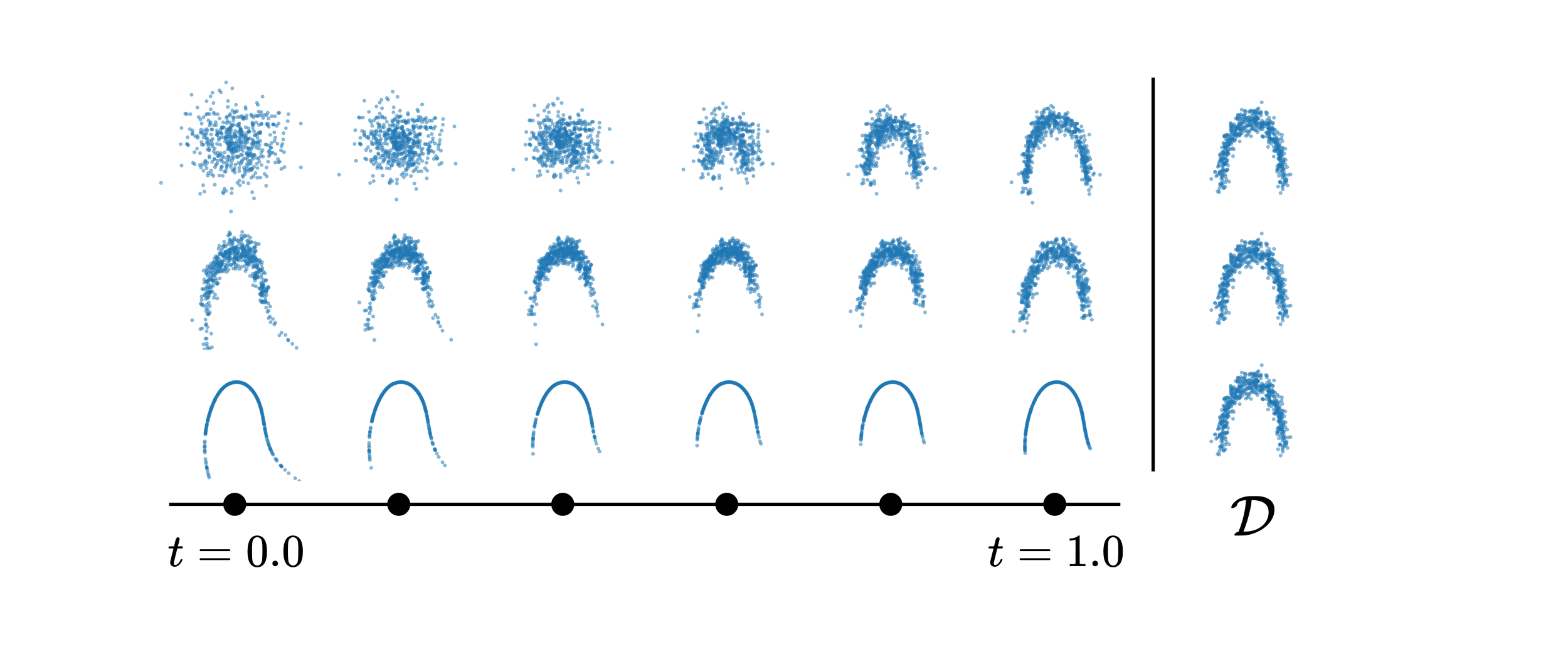}
        };
        \node[anchor=east] at (-5.0, 1.7) {\textbf{CFM}};
        \node[anchor=east] at (-5.0, 0.3) {\textbf{PFM}};
        \node[anchor=east] at (-5.0, -1.1) {\textbf{$1$-PFM}};
    \end{tikzpicture}
    \caption{Trajectories of \ac{CNF} (left) trained with \ac{CFM}, \ac{PFM} and $1$-\acs{PFM} objectives on the ARCH dataset compared to the data manifold $\mathcal{D}$ (right). At $t=0$ the trajectory starts with a standard normal distribution in the data space for \acs{CFM} and latent submanifold for ($1$-)\acs{PFM} mapped back to the data space.}
    \label{fig:generation arch figure}
\end{figure*}


%% file: sections/3_pullback_flow_matching.tex
We propose \textit{\acf{PFM}}, a novel framework for generative modeling on data manifolds using \textit{pullback geometry}. Our goal is to transform samples from a simple distribution $\boldsymbol{x}_0 \sim p$ on the data manifold $\mathcal{D}$ into a complex target distribution $\boldsymbol{x}_1 \sim q$, also on $\mathcal{D}$. Ideally, we would perform this transformation using \textit{\acf{RFM}}, see \autoref{sec:appendix background} for a summary, on $\big(\mathcal{D},(\cdot,\cdot)^{\mathcal{D}}\big)$ by optimizing the objective from \cite{chen2024flow},
\begin{align}
    &\mathcal{L}_{RFM}(\boldsymbol{\eta}) = \notag \\ & \E_{t,q(\boldsymbol{x}_1),p(\boldsymbol{x}_0)} 
    \Bigl( \bigl\| v_t\big(\gamma^{\mathcal{D}}_{\boldsymbol{x}_1,\boldsymbol{x}_0}(t);\boldsymbol{\eta}\big) 
    - \dot{\gamma}^{\mathcal{D}}_{\boldsymbol{x}_1,\boldsymbol{x}_0}(t) \bigr\|_{\gamma^{\mathcal{D}}_{\boldsymbol{x}_1,\boldsymbol{x}_0}(t)}^{\mathcal{D}} \Bigr)^2,
\end{align}
where $\boldsymbol{\eta}$ represents the learnable parameters of the parameterized vector field $v_t(\boldsymbol{x};\boldsymbol{\eta})$. Solving this objective on data manifolds becomes intractable as the training of \ac{RFM} is no longer simulation-free \citep{chen2024flow}. Existing methods address this challenge by employing restrictive and computationally intensive manifold mappings \citep{kapusniak2024metric}. We overcome this limitation by defining a new metric on the ambient space $\R^d$ using the \textit{pullback metric} \citep{diepeveen2024pulling} and assume a learned \textit{isometry} $\varphi_{\boldsymbol{\theta}}$ that approximates geodesics $\gamma^{\varphi_{\boldsymbol{\theta}}}$ on $\big(\R^d, (\cdot,\cdot)^{\varphi_{\boldsymbol{\theta}}}\big)$ to those $\gamma^{\mathcal{D}}$ on $\big(\mathcal{D},(\cdot,\cdot)^{\mathcal{D}}\big)$. Rewriting the \ac{RFM} objective under the pullback framework yields the objective,
\begin{align}
    &\mathcal{L}_{PFM}(\boldsymbol{\eta}) = \notag \\ & \E_{t,q(\boldsymbol{x}_1),p(\boldsymbol{x}_0)} \Bigl(\left\| v_t\Big(\gamma^{\varphi_{\boldsymbol{\theta}}}_{\boldsymbol{x}_1,\boldsymbol{x}_0}(t);\boldsymbol{\eta}\Big) - \dot{\gamma}^{\varphi_{\boldsymbol{\theta}}}_{\boldsymbol{x}_1,\boldsymbol{x}_0}(t) \right\|_{\gamma^{\varphi_{\boldsymbol{\theta}}}_{\boldsymbol{x}_1,\boldsymbol{x}_0}(t)}^{\varphi_{\boldsymbol{\theta}}}\Bigr)^2,
\end{align}
By applying \autoref{eq:pullback geodesic equation}, we reformulate the \ac{PFM} objective in terms of manifold mappings on $\mathcal{M}$,
\begin{align}
    & \mathcal{L}_{PFM}(\boldsymbol{\eta}) = \notag \\
    & \E_{t,q(\boldsymbol{x}_1),p(\boldsymbol{x}_0)} \Bigl(\left\| v_t\Big(\gamma^{\mathcal{M}}_{\varphi_{\boldsymbol{\theta}}(\boldsymbol{x}_1),\varphi_{\boldsymbol{\theta}}(\boldsymbol{x}_0)}(t);\boldsymbol{\eta}\Big) - \notag \right. \\ & \left. \dot{\gamma}^{\mathcal{M}}_{\varphi_{\boldsymbol{\theta}}(\boldsymbol{x}_1),\varphi_{\boldsymbol{\theta}}(\boldsymbol{x}_0)}(t) \right\|_{\gamma^{\mathcal{M}}_{\varphi_{\boldsymbol{\theta}}(\boldsymbol{x}_1),\varphi_{\boldsymbol{\theta}}(\boldsymbol{x}_0)}(t)}^{\mathcal{M}}\Bigr)^2,
\end{align}
Assuming a latent manifold $\mathcal{M}$ with closed-form mappings enables simulation-free training on data manifolds. For efficiency, we model the $d$-dimensional latent manifold as a product manifold, $\mathcal{M} = \mathcal{M}_{d'} \times \R^{d-d'}$. By encoding samples close to the submanifold $\mathcal{M}_{d'}\subset \mathcal{M}$, isometric learning ensures geodesics $\mathcal{M}_{d'}$ closely match geodesics on $\mathcal{M}$. As a result, we formulate the $d'$-\acs{PFM} objective,  
\begin{align}
    &\mathcal{L}_{d'-PFM}(\boldsymbol{\eta}) = \notag \\
    & \E_{t,q(\boldsymbol{x}_1),p(\boldsymbol{x}_0)} \Bigl(\left\| v_t\Big(\gamma^{\mathcal{M}_{d'}}_{\varphi_{\boldsymbol{\theta}}(\boldsymbol{x}_1),\varphi_{\boldsymbol{\theta}}(\boldsymbol{x}_0)}(t);\boldsymbol{\eta}\Big) - \notag \right. \\ &\left. \dot{\gamma}^{\mathcal{M}_{d'}}_{\varphi_{\boldsymbol{\theta}}(\boldsymbol{x}_1),\varphi_{\boldsymbol{\theta}}(\boldsymbol{x}_0)}(t)\right\|_{\gamma^{\mathcal{M}_{d'}}_{\varphi_{\boldsymbol{\theta}}(\boldsymbol{x}_1),\varphi_{\boldsymbol{\theta}}(\boldsymbol{x}_0)}(t)}^{\mathcal{M}_{d'}}\Bigr)^2,
\end{align}
The $d'$-\acs{PFM} objective offers two key benefits. Defining the objective on the submanifold $\mathcal{M}_{d'}$ results in computational speed-ups during training. Second, the known geometry on the submanifold simplifies the training dynamics of the vector field $v_t(\cdot;\boldsymbol{\eta})$, requiring fewer parameters $\boldsymbol{\eta}$ to learn the sampling trajectories of the data manifold, see \autoref{tab:comparison between generative methods for manifold distribution learning}.

%% file: sections/4_learning_isometries.tex
The motivation for learning isometries $\varphi_{\boldsymbol{\theta}}$—metric-preserving diffeomorphisms—is to obtain a latent (sub)manifold that supports interpolation with closed-form geometric mappings, facilitating efficient simulation-free training of \ac{PFM}. Our approach aligns with the isometry learning framework introduced by Diepeveen \cite{diepeveen2024pulling}, but differs in two crucial ways. First, we propose a more expressive parameterization of learnable diffeomorphisms via neural ordinary differential equations (NODEs). Second, we introduce a novel training objective that enables scalable isometric learning on data manifolds.

\subsection{Parameterizing Diffeomorphisms}
\label{subsec:parameterizing diffeomorphisms}
We parameterize diffeomorphisms, invertible and differentiable functions between two manifolds, specifically $\varphi: \R^d \rightarrow \mathcal{M}$. In practice, we construct a product manifold, $\mathcal{M} = \mathcal{M}_{d'} \times \R^{d-d'}$ and the diffeomorphism $\varphi$ as,
\begin{equation}
    \varphi : = [\psi^{-1},\boldsymbol{I}_{d-d'}] \circ \phi \circ T_{\boldsymbol{\mu}},
\end{equation}
where $\psi: U \rightarrow \R^{d'}$ a chart on a geodesically convex subset $U \subset \mathcal{M}_{d'}$ of the $d'$-dimensional latent submanifold $\big(\mathcal{M}_{d'},(\cdot,\cdot)_{\mathcal{M}_{d'}}\big)$. See \autoref{sec:appendix closed form manifold mappings} for a list of manifolds with closed-form expressions for the exponential and logarithmic maps, and how these maps can be used to construct a chart $\psi$. Furthermore, $\phi: \R^d \rightarrow \R^d$ is a diffeomorphism and $T_{\boldsymbol{\mu}}(\boldsymbol{x}) = \boldsymbol{x} - \boldsymbol{\mu}$, with $\boldsymbol{\mu}$ the average of the datapoints. We choose this construction because the manifold hypothesis translates to assuming the data manifold is homeomorphic to $\mathcal{M}_{d'}$. In such case, the rest of the latent manifold should be mapped close to zero, e.g. $\varphi(\boldsymbol{x}_i)$ is close to $\mathcal{M}_{d'} \times \boldsymbol{0}^{d-d'}$ in terms of the metric on $\mathcal{M}$.

We generate the diffeomorphism $\phi$ by learning a \ac{NODE} \citep{chen2018neural}. The advantage of this approach is threefold, \textit{i)} this parameterization of diffeomorphisms is more expressive and efficient to train compared to Invertible Residual Networks \citep{behrmann2019invertible} as chosen by \cite{diepeveen2024pulling}, \textit{ii)} based on some mild technical assumptions a \ac{NODE} can be proven to generate proper diffeomorphisms, see \autoref{sec:appendix neural odes parameterize diffeomorphisms} for the proof, and \textit{iii)} numerically the accuracy and invertibility of the generated flow can be controlled through smaller step-sizes and higher-order solvers. 

To define the diffeomorphism $\phi_{\boldsymbol{\theta}}: \mathbb{R}^d \rightarrow \mathbb{R}^d$, we start with the \ac{NODE} governing the flow:
\begin{equation}
    \frac{d\boldsymbol{z}(t)}{dt} = f(\boldsymbol{z}(t); \boldsymbol{\theta}),
\end{equation}
where $f:\R^d\rightarrow\R^d$ is a vector field parameterized by a \ac{MLP} with Swish activation functions and a sine-cosine time embedding and $\boldsymbol{\theta}$ denotes the parameters of the \ac{MLP}. Given an initial condition $\boldsymbol{z}(0) = \boldsymbol{x}$, the solution to this \ac{NODE} is:
\begin{equation}
    \phi_{\boldsymbol{\theta}}(\boldsymbol{x}) := \boldsymbol{x} + \int_{0}^{1} f(\boldsymbol{z}(t); \boldsymbol{\theta}) \, dt.
\end{equation}

To obtain the inverse $\phi_{\boldsymbol{\theta}}^{-1}$ one has to integrate the differential equation backwards in time with initial condition $\boldsymbol{z}(1)$. To solve the \ac{NODE} we implemented a Runge-Kutta solver in \texttt{JAX}, see \autoref{sec:appendix training procedure} for further architectural and training related details. 

\subsection{Learning Objective}
\label{subsec:learning objective}
The primary objectives of learning isometries are \textit{i)} to map the data manifold $\big( \mathcal{D}, (\cdot,\cdot)^{\mathcal{D}} \big)$ into a low-dimensional geodesic subspace of $\big(\mathcal{M},(\cdot,\cdot)^{\mathcal{M}} \big)$, specifically $\mathcal{M}_{d'} \subset \mathcal{M}$, and \textit{ii)} to preserve local isometry, as motivated by Proposition 2.1 and Theorems 3.4, 3.6, and 3.8 from \cite{diepeveen2024pulling}.

We improve the training objective of \citet{diepeveen2024pulling} by incorporating \ref{eq:graph matching loss} for isometric learning, enforcing global isometry between data and latent manifolds \citep{zhu2014matrix}, ensuring each sample remains equally isometric to all others. Additionally, we use \ref{eq:metric preserving loss} and \ref{eq:submanifold loss my method} to map the data manifold $\mathcal{D}$ onto the lower-dimensional geodesic subspace $\mathcal{M}_{d'}$.

The original objective enforces local isometry—preserving geodesic distances in small neighborhoods—via the pullback metric's Riemannian tensor $(\cdot,\cdot)^{\varphi}$. However, this is computationally intractable and poorly scalable. We address this by using the regularization in \ref{eq:stability regularization} from \cite{finlay2020train}, which more efficiently enforces local isometry, leading to a scalable objective,
\begin{align}
    \mathcal{L}(\boldsymbol{\theta}) &= \alpha_{1}\frac{1}{n^2}\sum^{n}_{i=1} \sum^{n}_{j=1} \|d^{\varphi_{\boldsymbol{\theta}}}_{\mathbb{R}^d}(\boldsymbol{x}_i,\boldsymbol{x}_j) - d_{i,j} \|^2 \tag{\textcolor{kth-blue}{\textbf{global isometry loss}}} \label{eq:metric preserving loss}\\
    &+ \alpha_{2} \frac{1}{n}\sum^{n}_{i=1}\sum_{j\neq i} \| (\boldsymbol{d}^{\varphi_{\boldsymbol{\theta}}}_{\mathbb{R}^d}(\boldsymbol{x}_i,\boldsymbol{x}_{\cdot}) - \boldsymbol{d}^{\varphi_{\boldsymbol{\theta}}}_{\mathbb{R}^d}(\boldsymbol{x}_j,\boldsymbol{x}_{\cdot})) \notag \\ & \quad - (\boldsymbol{d}_{i,\cdot} -\boldsymbol{d}_{j,\cdot})\|^2 \tag{\textcolor{kth-blue}{\textbf{graph matching loss}}} \label{eq:graph matching loss}\\
    & + \alpha_{3}\frac{1}{n} \sum_{i=1}^n \left\| \begin{bmatrix}
                \boldsymbol{0}_{d'}  & \emptyset\\
                \emptyset & \boldsymbol{I}_{d-d'} 
            \end{bmatrix} (\phi_{\boldsymbol{\theta}} \circ T_{\boldsymbol{\mu}})(\boldsymbol{x}_i) \right\|_1 \tag{\textcolor{kth-blue}{\textbf{submanifold loss}}} \label{eq:submanifold loss my method}\\
    &+ \alpha_{4} \frac{1}{n} \sum_{i=1}^n \int_0^1 \| \boldsymbol{\varepsilon}^T\nabla f_{\boldsymbol{\theta}}(\boldsymbol{z}_i(t))\|^2 \; dt. \tag{\textcolor{kth-blue}{\textbf{stability regularization}}} \label{eq:stability regularization}
\end{align}
Here, $\boldsymbol{\varepsilon} \sim \mathcal{N}(\boldsymbol{0},\boldsymbol{I})$ and $\boldsymbol{d}^{\varphi_{\boldsymbol{\theta}}}_{\mathbb{R}^d}(\boldsymbol{x}_i,\boldsymbol{x}_{\cdot})$ and $\boldsymbol{d}_{i,\cdot}$ denote the columns of the distance matrices induced by $(\cdot,\cdot)^{\varphi}$ and $(\cdot,\cdot)^{\mathcal{D}}$. The benefit of this formulation is that it only requires approximating geodesic distances $d_{i,j}$ on the data manifold $\mathcal{D}$, without needing to calculate or differentiate the metric tensor. In \autoref{sec:experiments}, we demonstrate the effectiveness of the graph matching loss and stability regularization through an ablation study on synthetic and high-dimensional protein dynamics trajectories. We do not include an ablation of the global isometry loss and submanifold losses, as these have been thoroughly examined in \citep{diepeveen2024pulling}, and our experiments showed consistent results with those previously reported.

\begin{table*}[t!]
    \centering
    \small
    \caption{Ablation study of isometric learning for \acs{ARCH} dataset and \acs{I-FABP} protein dynamics datasets for \ref{eq:graph matching loss} (GM) and \ref{eq:stability regularization} (Stability). In both cases we choose $\mathcal{M}_{d'}=\R$. We report the means for invertibility ($\downarrow$), low-dimensionality ($\downarrow$) and isometry ($\downarrow$) with standard devations denoted by $\pm$. The distance $(\cdot,\cdot)^{\mathcal{D}}$ we assume on the data manifold $\mathcal{D}$ is a locally Euclidean distance based on Isomap \citep{tenenbaum2000global}.}
    \label{tab:ablation_study}
    \begin{tabular}{clcccc}
            \toprule
            \textbf{Data} & \textbf{Metric} & \textbf{None} & \textbf{GM} & \textbf{Stability} & \textbf{Both} \\
            \midrule
            \multirow{6}{*}{\centering \raisebox{-0.2cm}[0pt][0pt]{\rotatebox{90}{\textbf{\acs{ARCH}}}}}
            & \multirow{2}{*}{Invertibility} & $7.637 \cdot 10^{-1}$ & $3.585 \cdot 10^{-2}$ & $\boldsymbol{8.198 \cdot 10^{-5}}$ & $1.011 \cdot 10^{-4}$ \\[-0.2cm]
            & & \textsubscript{$\pm 9.872 \cdot 10^{-1}$} & \textsubscript{$\pm 1.939 \cdot 10^{-2}$} & \textsubscript{$\pm 1.061 \cdot 10^{-5}$} & \textsubscript{$\pm 6.069 \cdot 10^{-5}$} \\[0.1cm]
            & \multirow{2}{*}{Low-Dimensionality} & $6.520 \cdot 10^{-4}$ & $\boldsymbol{4.531 \cdot 10^{-4}}$ & $1.407 \cdot 10^{-2}$ & $1.373 \cdot 10^{-2}$ \\[-0.2cm]
            & & \textsubscript{$\pm 9.521 \cdot 10^{-5}$} & \textsubscript{$\pm 3.341 \cdot 10^{-5}$} & \textsubscript{$\pm 8.414 \cdot 10^{-4}$} & \textsubscript{$\pm 6.768 \cdot 10^{-4}$} \\[0.1cm]
            & \multirow{2}{*}{Isometry} & $2.334 \cdot 10^{-3}$ & $\boldsymbol{1.464 \cdot 10^{-3}}$ & $2.018 \cdot 10^{-3}$ & $1.544 \cdot 10^{-3}$ \\[-0.2cm]
            & & \textsubscript{$\pm 1.466 \cdot 10^{-4}$} & \textsubscript{$\pm 1.221 \cdot 10^{-4}$} & \textsubscript{$\pm 5.791 \cdot 10^{-5}$} & \textsubscript{$\pm 2.025 \cdot 10^{-4}$} \\[0.1cm]
            \midrule
            \multirow{6}{*}{\centering \raisebox{-0.2cm}[0pt][0pt]{\rotatebox{90}{\textbf{\acs{I-FABP}}}}} 
            & \multirow{2}{*}{Invertibility} 
            & $2.995 \cdot 10^{-5}$ & $2.891 \cdot 10^{-5}$ & $2.973 \cdot 10^{-5}$ & $\boldsymbol{2.809 \cdot 10^{-5}}$ \\[-0.2cm]
            & & \textsubscript{$\pm 8.945 \cdot 10^{-6}$} & \textsubscript{$\pm 4.968 \cdot 10^{-6}$} & \textsubscript{$\pm 7.560 \cdot 10^{-6}$} & \textsubscript{$\pm 8.982 \cdot 10^{-6}$} \\[0.1cm]
            & \multirow{2}{*}{Low-Dimensionality} & $\boldsymbol{1.378 \cdot 10^{-1}}$ & $1.379 \cdot 10^{-1}$ & $1.379 \cdot 10^{-1}$ & $\boldsymbol{1.378 \cdot 10^{-1}}$ \\[-0.2cm]
            & & \textsubscript{$\pm 1.952 \cdot 10^{-4}$} & \textsubscript{$\pm 1.424 \cdot 10^{-4}$} & \textsubscript{$\pm 1.788 \cdot 10^{-4}$} & \textsubscript{$\pm 1.981 \cdot 10^{-4}$} \\[0.1cm]
            & \multirow{2}{*}{Isometry} & $2.889 \cdot 10^{-3}$ & $2.898 \cdot 10^{-3}$ & $2.919 \cdot 10^{-3}$ & $\boldsymbol{2.887 \cdot 10^{-3}}$ \\[-0.2cm]
            & & \textsubscript{$\pm 1.384 \cdot 10^{-4}$} & \textsubscript{$\pm 1.667 \cdot 10^{-4}$} & \textsubscript{$\pm 1.571 \cdot 10^{-4}$} & \textsubscript{$\pm 1.387 \cdot 10^{-4}$} \\
            \bottomrule
    \end{tabular}
\end{table*}

%% file: sections/5_experiments.tex
The goal of this paper is to learn interpolatable latent (sub)manifolds for generation on data manifolds. We achieve this through isometric learning in the framework of pullback geometry. In this section we validate our methods on synthetic, simulated and experimental datasets, for full descriptions see \autoref{sec:appendix data description}. For details on the training procedure and hyperparameter settings we refer the reader to \autoref{sec:appendix training procedure}.

We begin our experiments with an ablation study of \ref{eq:graph matching loss} and \ref{eq:stability regularization}, demonstrating the benefits of including both terms for learning isometries. Second, we compare (latent) interpolation methods with interpolation on the latent manifold $\mathcal{M}$, $(\cdot,\cdot)^{\mathcal{M}}$-interpolation, and on the latent submanifold $\mathcal{M}_{d'}$, $(\cdot,\cdot)^{\mathcal{M}_{d'}}$-interpolation. We demonstrate that we can accurately interpolate on the data manifold by interpolating on the latent (sub)manifold \footnote{In these experiments we do not report interpolation through the \ac{RAE} by \cite{diepeveen2024pulling} due to the intractability of the training objective for the higher-dimensional datasets.}. Third, we validate \ac{PFM} as a generative model on data manifolds and discuss how sample generation is improved by generating on the submanifold $\mathcal{M}_{d'}$. Finally, we inspect the designability of the latent manifold through the choice of metric $(\cdot,\cdot)^{\mathcal{D}}$ in the task of small protein design. 

\subsection{Ablation Study}
\label{subsec:ablation study}

The goal of the ablation study is to evaluate the effectiveness of the reformulated objective function for learning isometries. To this end, we perform an ablation study for both the \ref{eq:graph matching loss} and \ref{eq:stability regularization} on a synthetic \acs{ARCH} dataset ($n=500$, $d=2$) in the spirit of \cite{tong2020trajectorynet} and a coarse-grained protein dynamics datasets of \ac{I-FABP} ($n=500$, $d=131 \times 3$). We report three metrics on the validation set of 20 \% of the data, invertibility $\varepsilon_{inv} = \frac{1}{n} \sum_{i=1}^n \|\boldsymbol{x}_i - \varphi_{\boldsymbol{\theta}}^{-1}\big(\varphi_{\boldsymbol{\theta}}(\boldsymbol{x}_i)\big)\|^2$, low-dimensionality $\varepsilon_{ld} = \frac{1}{n} \sum_{i=1}^n \left \| \begin{bmatrix}
                \boldsymbol{0}_{d'} & \emptyset\\
                \emptyset & \boldsymbol{I}_{d-d'}  
            \end{bmatrix}\phi_{\boldsymbol{\theta}}(\boldsymbol{x}_i) \right \|^2_1$ and isometry $\varepsilon_{iso} = \frac{1}{n^2} \sum^n_{i=1} \sum^n_{j=1} \|d_{i,j}-d_{\mathcal{M}}\big(\varphi(\boldsymbol{x}_i),
        \varphi(\boldsymbol{x}_j)\big)\|^2$.

\textbf{Result.} \autoref{tab:ablation_study} demonstrates that incorporating both the graph matching loss and stability regularization improves the invertibility and isometry metrics across both datasets, with the combined approach yielding both a low $\varepsilon_{inv}$ and $\varepsilon_{iso}$ values, indicating enhanced model performance in preserving the geometry of the data in the synthetic dataset as well as the more noisy and high dimensional dataset.

\begin{figure}[h!]
    \centering
    \includegraphics[trim = 75 20 75 20,clip,width = 0.8\linewidth]{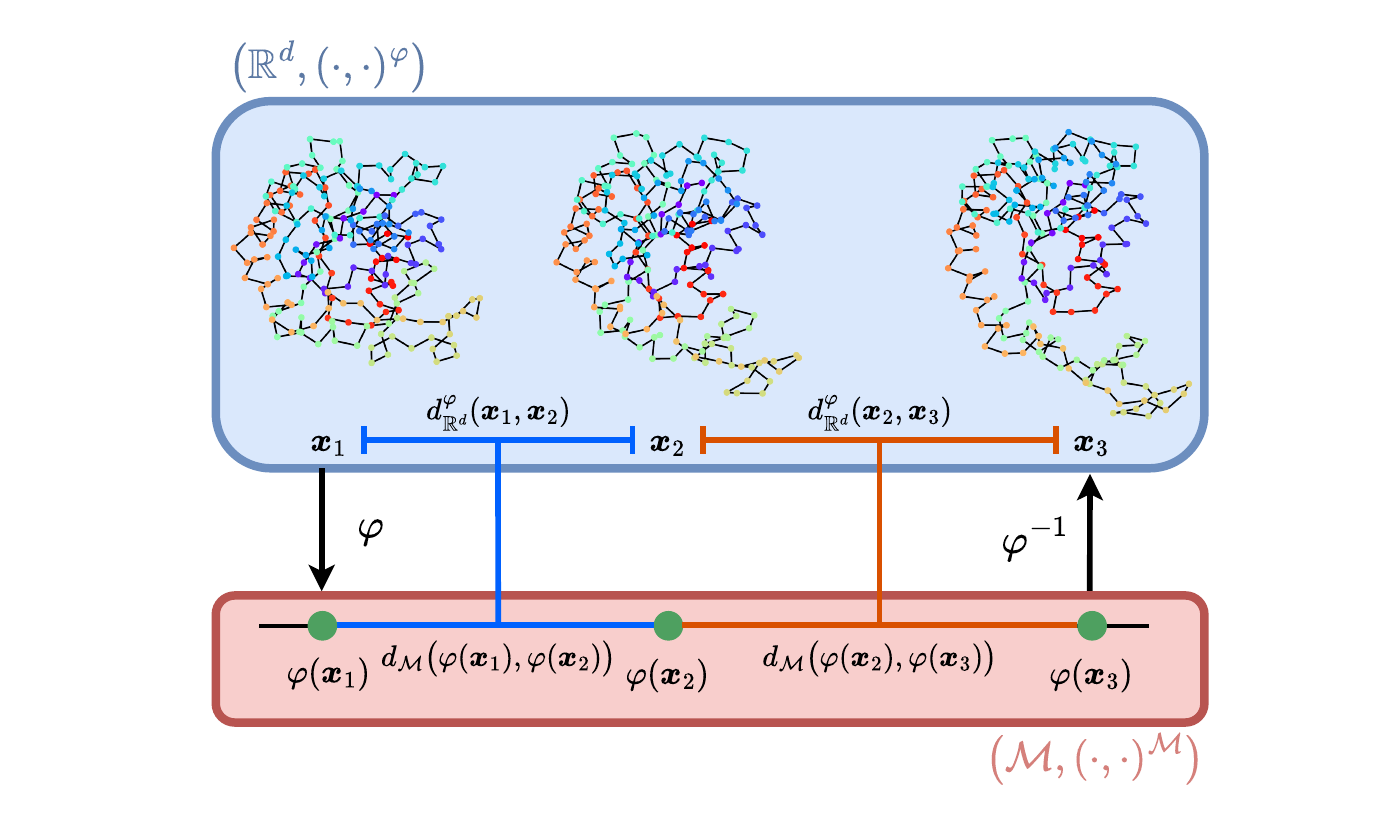}
    \caption{Isometric learning for \acs{AK} dataset, coarse-grained protein conformation data. We define a new metric $(\cdot,\cdot)^{\varphi}$ on the ambient space, $\mathbb{R}^d$ ($d=214 \times 3$), by learning a diffeomorphism $\varphi: \R^{d} \rightarrow \mathcal{M}$ that preserves a locally Euclidean metric $(\cdot,\cdot)^{\mathcal{D}}$ on the latent manifold $\mathcal{M}=\mathcal{M}_{d'}\times \R^{d-d'}$ for $d'=1$.}
    \label{fig:explainer for AK isometric learning}
\end{figure}

\subsection{Interpolation Experiments}
\label{subsec:interpolation experiments}
The goal of isometric learning is to learn an interpolatable latent (sub)manifold of the data manifold with closed-form manifold mappings. To evaluate whether interpolation on the latent (sub)manifold accurately reflects interpolation on the data manifold, we conduct an interpolation experiment using the synthetic \acs{ARCH} dataset, as well as the molecular dynamics datasets of \ac{AK} and \ac{I-FABP}, see \autoref{fig:explainer for AK isometric learning} for the example of isometric learning on the \acs{AK} dataset. 

\begin{figure}[h!]
    \centering
    \includegraphics[trim=10 10 10 10,clip,width=0.6\linewidth]{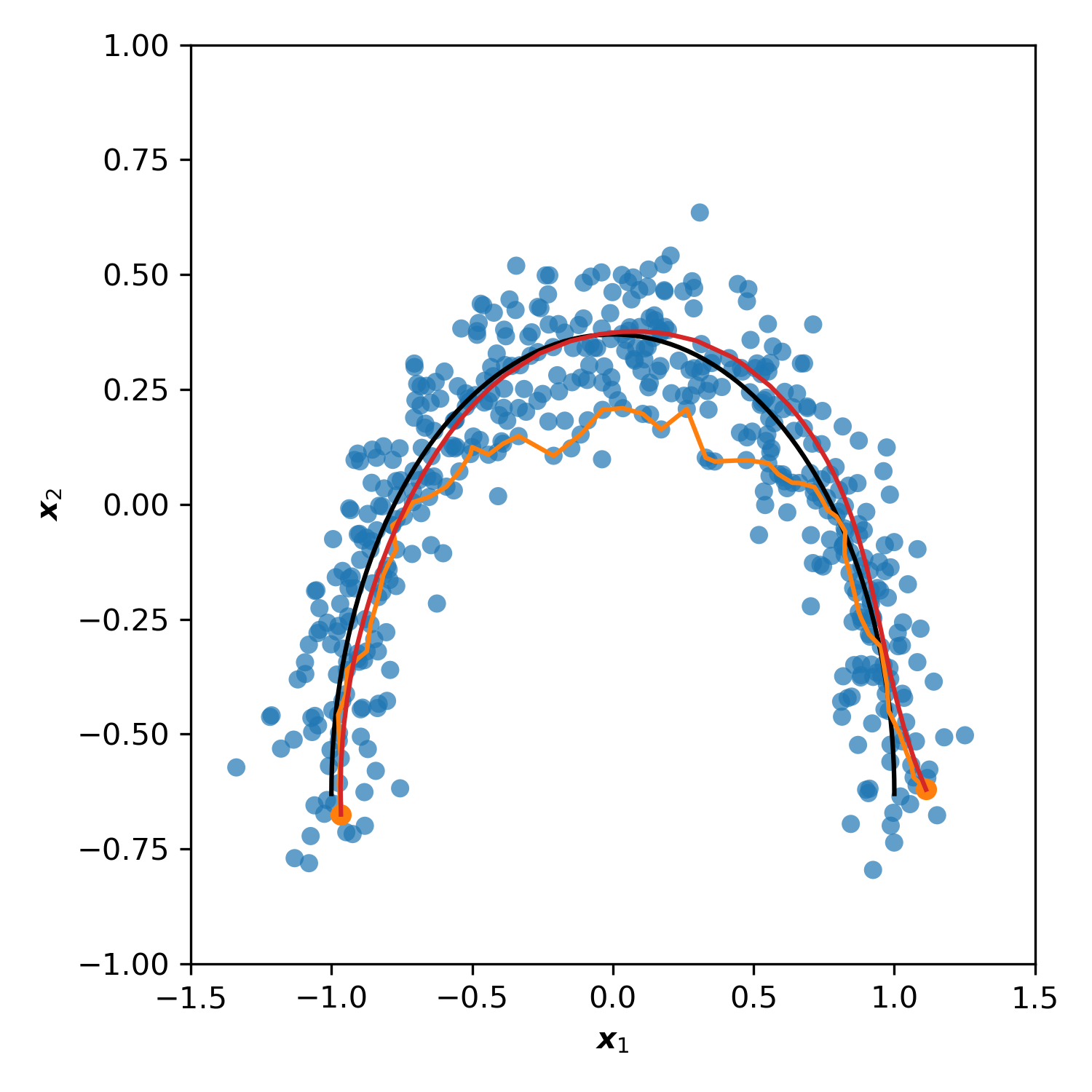}
    \captionof{figure}{Example of $(\cdot,\cdot)^{\mathcal{M}_{d'}}$-interpolation for \acs{ARCH} dataset in \textcolor{tab_red}{\textbf{red}}. In \textcolor{tab_blue}{\textbf{blue}} the dataset $\{\boldsymbol{x}_i\}^n_{i=1}$, \textcolor{black}{\textbf{black}} the true submanifold $\mathcal{M}_{d'}$, the half circle, and in \textcolor{tab_orange}{\textbf{orange}} the Isomap geodesic between \textcolor{tab_orange}{\textbf{orange points}}.}
    \label{fig:geodesic example figure}
\end{figure}

In both cases we choose $\mathcal{M}_{d'}=\R$, see \autoref{sec:appendix manifold and metric selection} for guidance on latent manifold and metric selection. We approximate the metric on the data manifold $(\cdot,\cdot)^{\mathcal{D}}$ through the length of Isomap's geodesics \cite{tenenbaum2000global}, see \autoref{fig:geodesic example figure} for an example. We compare the accuracy of the $100$ longest geodesics between points in the test set for multiple (latent) interpolation methods. 

\begin{table*}
    \centering
    \small
    \caption{\Ac{RMSE} ($\downarrow$) of the $100$ longest isomap geodesics in the test set for $3$ different seeds for linear interpolation, $(\cdot,\cdot)^{\mathcal{M}}$-interpolation and $(\cdot,\cdot)^{\mathcal{M}_{d'}}$-interpolation, compared to latent interpolation methods \acp{VAE} \citep{kingma2013auto}, $\beta-$VAEs \citep{higgins2017beta} and GRAE \citep{duque2022geometry}.}   
    \label{tab:results}
    \begin{tabular}{lccccc}
        \toprule
        \textbf{Interpolation} & \textbf{Latent} & \textbf{\acs{ARCH}} & \textbf{Swiss Roll} & \textbf{\acs{AK}} & \textbf{\acs{I-FABP}} \\ 
        \midrule 
        Linear & \textcolor{tab_red}{\xmark} & $0.331_{\pm 0.049}$ & $0.573_{\pm 0.018}$ & $0.554_{\pm 0.131}$ & $0.494_{\pm 0.022}$ \\
        \acs{VAE} & \textcolor{tab_green}{\cmark} & $0.526_{\pm 0.024}$ & $0.596_{\pm 0.085}$ & $1.235_{\pm 0.477}$ & $0.405_{\pm 0.023}$ \\
        $\beta$-\acs{VAE} & \textcolor{tab_green}{\cmark} & $0.527_{\pm 0.025}$ & $0.640_{\pm 0.066}$ & $0.919_{\pm 0.631}$ & $0.368_{\pm 0.009}$ \\
        GRAE (Isomap) & \textcolor{tab_green}{\cmark} & $0.426_{\pm 0.076}$ & $0.568_{\pm 0.024}$ & $2.030_{\pm 0.579}$ & $0.442_{\pm 0.005}$ \\
        GRAE (PHATE) & \textcolor{tab_green}{\cmark} & $\boldsymbol{0.128}_{\pm 0.052}$ & $0.660_{\pm 0.150}$ & $1.012_{\pm 0.395}$ & $0.474_{\pm 0.040}$ \\
        $(\cdot,\cdot)^{\mathcal{M}}$ & \textcolor{tab_green}{\cmark} & $\boldsymbol{0.097}_{\pm 0.030}$ & $\boldsymbol{0.159}_{\pm 0.054}$ & $0.296_{\pm 0.058}$ & $0.415_{\pm 0.025}$ \\
        $(\cdot,\cdot)^{\mathcal{M}_{d'}}$ & \textcolor{tab_green}{\cmark} & $\boldsymbol{0.109}_{\pm 0.026}$ & $\boldsymbol{0.159}_{\pm 0.055}$ & $\boldsymbol{0.219}_{\pm 0.012}$ & $\boldsymbol{0.292}_{\pm 0.006}$ \\
        \bottomrule
    \end{tabular}
\end{table*}

\textbf{Result.} The $(\cdot,\cdot)^{\mathcal{M}_{d'}}$- and $(\cdot,\cdot)^{\mathcal{M}}$-interpolation achieves superior interpolation accuracy with lower \acs{RMSE} variablity compared to other models, indicating more robust and reliable interpolation. $(\cdot,\cdot)^{\mathcal{M}_{d'}}$-interpolation specifically demonstrates improvements over other methods in the more stochastic and seemingly higher dimensional \ac{AK} ($d=639$) and \ac{I-FABP} ($d=642$) datasets. This improvement suggests that compressing the latent representation into a lower-dimensional space reduces the noise while accurately capturing the underlying data manifold. Our findings demonstrate that accurate interpolation of protein dynamics trajectories of \ac{AK} and \ac{I-FABP} can be achieved using a single-dimensional latent manifold. This method shows promise for improving protein dynamics simulations.

\subsection{Generation Experiments}
\label{subsec:generation experiments}

We demonstrate the effectiveness of our proposed method \ac{PFM} for generation on data manifolds $\mathcal{D}$. We train two \acp{PFM}, one using the latent manifold $\mathcal{M}$ and one using the lower dimensional latent submanifold $\mathcal{M}_{d'}$, named \ac{PFM} and $d'$-\acs{PFM} respectively. Additionally, we train a \ac{CFM} model on the raw data as a comparison. A visual example of the learned generative flows over time for the \acs{ARCH} dataset can be viewed in \autoref{fig:generation arch figure}. To evaluate our generative methods, we use the $1$-\ac{NN} accuracy \citep{lopez2016revisiting}, which measures how well the generated point clouds match the reference point clouds. Each point cloud is classified by finding its nearest neighbor in the combined set of generated and reference point clouds. The accuracy reflects how similar the generated point clouds are to the reference set, with an accuracy close to $50\%$ indicating successful learning of the target distribution.

\begin{table*}[bp]
    \centering
    \small
    \caption{Evaluation of generative model performance across dimensionality of (latent) (sub)manifold ($\downarrow$), number of model parameters, denoted by \# pars ($\downarrow$), and $1$-\acs{NN} accuracy ($1$-\acs{NN}$\rightarrow 0.5$). The $1$-\acs{NN} metric measures the generative quality, with values closer to $0.5$ indicating better performance.}
    \label{tab:comparison between generative methods for manifold distribution learning}
    \begin{tabular}{lcccccc}
        \toprule
        & \multicolumn{3}{c}{\textbf{\acs{ARCH}}} & \multicolumn{3}{c}{\textbf{Swiss}}\\
        \cmidrule(lr){2-4} \cmidrule(lr){5-7} 
        \textbf{Model} & $\boldsymbol{\dim}$ & \textbf{\# pars} & $\boldsymbol{1}$\textbf{-\acs{NN}} & $\boldsymbol{\dim}$ & \textbf{\# pars} & $\boldsymbol{1}$\textbf{-\acs{NN}}\\
        \midrule 
        \acs{CFM} & $2$ & $50562$ & $0.295_{\pm 0.031}$ & $2$ & $50691$ & $0.870_{\pm 0.016}$\\
        \acs{PFM} & $2$ & $50562$ & $0.262_{\pm 0.025}$ & $2$ & $50691$ & $0.795_{\pm 0.011}$\\
        $1$-\acs{PFM} & $\boldsymbol{1}$ & $\boldsymbol{5697}$ & $\boldsymbol{0.487}_{\pm 0.027}$ & $\boldsymbol{1}$ & $\boldsymbol{16066}$ & $\boldsymbol{0.789}_{\pm 0.019}$\\
        \midrule 
        & \multicolumn{3}{c}{\textbf{\acs{AK}}} & \multicolumn{3}{c}{\textbf{\acs{I-FABP}}}\\
        \cmidrule(lr){2-4} \cmidrule(lr){5-7} 
        \textbf{Model} & $\boldsymbol{\dim}$ & \textbf{\# pars} & $\boldsymbol{1}$\textbf{-\acs{NN}} & $\boldsymbol{\dim}$ & \textbf{\# pars} & $\boldsymbol{1}$\textbf{-\acs{NN}}\\
        \acs{CFM} & $642$ & $4682325$ & $0.386_{\pm 0.000}$ & $393$ & $1789941$ & $0.365_{\pm 0.004}$\\
        \acs{PFM} & $642$ & $4682325$ & $0.356_{\pm 0.097}$ & $393$ & $1789941$ & $0.452_{\pm 0.017}$\\
        $1$-\acs{PFM} & $\boldsymbol{1}$ & $\boldsymbol{5697}$ & $\boldsymbol{0.464}_{\pm 0.022}$ & $\boldsymbol{1}$ & $\boldsymbol{5697}$ & $\boldsymbol{0.508}_{\pm 0.006}$\\
        \bottomrule
    \end{tabular}
\end{table*}

\textbf{Result.} \autoref{fig:generation arch figure} we see that the learned isometry to the latent manifold $\mathcal{M}$ acts as a strong manifold prior, capturing the manifold structure at the start of the \ac{CNF} trajectory ($t=0.0$). Additionally, the learned isometry to the latent submanifold $\mathcal{M}_{d'}$ captures the noiseless manifold revealing the underlying manifold used to generate the data. Through this strong (noiseless) manifold prior, we see that both \ac{PFM} and $1$-\acs{PFM} approximate the distribution on the manifold earlier in the trajectory and better. \autoref{tab:comparison between generative methods for manifold distribution learning} highlights the effectiveness of the $1$-\acs{PFM} model in generative tasks. The $1$-\acs{PFM} model leverages the lower-dimensional isometric latent manifold $\mathcal{M}_{d'}$, significantly reducing the number of parameters required. Training the $1$-\acs{PFM} is significantly faster due to the reduction in parameters and the dimensionality of the training samples. The $1$-\acs{NN} accuracy for $1$-\acs{PFM} approaches the ideal $0.5$ across all datasets, indicating that this model better captures the underlying distribution on the data manifold compared to CFM and \ac{PFM}.

\subsection{Designable Latent Manifolds for Novel Protein Engineering}
\label{subsec:designable latent spaces for protein design}

The goal of these experiments is to design a latent manifold that captures biologically relevant properties of protein sequences, enabling the generation of novel proteins with specific characteristics. By leveraging our method's flexibility in defining the metric on the data manifold $(\cdot, \cdot)^{\mathcal{D}}$, we structure the latent space such that it captures protein properties, such as sequence similarity, hydrophobicity, hydrophobic moment, charge, and isoelectric point.

To achieve this, we use protein sequences of up to 25 amino acids from the \ac{GRAMPA} dataset (see \autoref{sec:appendix data description} for details). We construct the following custom metric on the data manifold,
\begin{align}
    d_{\mathcal{D}}(x_i,x_j) &= d_{\text{Levenshtein}}(x_i,x_j) + d_{\text{hydrophobicity}}(x_i,x_j) \\
    &+ d_{\text{hydrophobic moment}}(x_i,x_j) + d_{\text{charge}}(x_i,x_j) \\
    &+ d_{\text{isoelectric point}}(x_i,x_j),
\end{align}
where the Levenshtein distance measures the number of single-character edits (insertions, deletions, or substitutions) required to transform one sequence into another.

For the remaining four properties---hydrophobicity, hydrophobic moment, charge, and isoelectric point---distances are computed using the difference in property values between sequences. We define the (pseudo)distance as,
\begin{equation}
    d_{\text{[property]}}(x_i,x_j) = |f_{\text{property}}(x_i) - f_{\text{property}}(x_j)|.
\end{equation}
These (pseudo)distances are standardized by dividing by the maximum observed distance in the training data. Since the Levenshtein distance is a proper metric, we ensure that the combined distance $d_{\mathcal{D}}(x_i,x_j)$ remains a valid distance metric. We use the designed metric $(\cdot,\cdot)^{\mathcal{D}}$ on the space of protein sequences with at most 25 amino acids in the \ac{GRAMPA} dataset to learn an isometry that preserves this metric on the latent manifold $\mathcal{M}$. 

To generate protein sequences with specific properties, we sample from a normal distribution around the data points in the latent manifold $\boldsymbol{z} \in \mathcal{M}$. The variability of this sampling process is aligned with the latent variability of the training data $\sigma_{\boldsymbol{z}_{train}}$, scaled by a temperature factor $\tau$, resulting in the following expression,
\begin{align}
    \boldsymbol{z}^{(analogue)}_i &= \boldsymbol{z}_i + \tau \mathcal{N}(\boldsymbol{0}, \sigma_{\boldsymbol{z}_{train}}\boldsymbol{I}), \text{ and} \\
    \boldsymbol{x}^{(analogue)}_i &= \varphi_{\boldsymbol{\theta }}^{-1}(\boldsymbol{z}_i) \text{ for } i = 1, \ldots, n_{test}.
\end{align}
This sampling methodology is referred to as \textit{analogue generation}, as it does not involve explicitly learning the distribution over the latent manifold. Instead, it generates novel sequences by sampling around existing data points on the latent (sub)manifold of the test set. 

To evaluate the effectiveness of the generated sequences, we measure the number of unique sequences that were not present in the original dataset and compare the properties of the generated samples to the properties of their base points. For further specifics on hyperparameters and training procedures, refer to \autoref{sec:appendix training procedure}. 

\begin{table}[h!]
    \centering
    \small
    \caption{Unique protein sequences generated via analogue generation on the latent manifold $\mathcal{M}$ at various temperatures ($\tau$). The table presents the total sequences generated (Total), those already in the dataset (In Data), and the number of novel sequences (Novel). We perform a \ac{KS} test at a $5\%$ significance level to compare novel sequences with their base points. Non-significant \ac{KS} values are shown as X/Y, where X is the number of non-significant properties and Y is the total properties tested.}
    \label{tab:unique_samples_across_temperatures_novel_focus}
    \begin{tabular}{lcccc}
        \toprule
       $\boldsymbol{\tau}$ & \textbf{Total} & \textbf{In Data} & \textbf{Novel} & \textbf{Non-Sign. \acs{KS}} \\
        \midrule
        0.01 & 689 & 652 &  37 & 5/5\\
        0.05 & 689 & 103 & 586 & 5/5\\
        0.1  & 689 &  35 & 654 & 5/5\\
        0.2  & 689 &  12 & 677 & 2/5\\
        0.5  & 689 &   1 & 688 & 1/5\\
        1    & 689 &   0 & 689 & 0/5\\
        \bottomrule
    \end{tabular}
\end{table}

\textbf{Results.} Designed latent manifolds facilitated the generation of diverse novel protein sequences, demonstrating the effectiveness of our analogue generation methodology. As shown in Table \ref{tab:unique_samples_across_temperatures_novel_focus}, increasing the temperature parameter (\(\tau\)) directly influenced the diversity of generated sequences. At lower temperatures (\(\tau \leq 0.1\)), many unique sequences emerged while maintaining similarity to their base points, as indicated by non-significant KS test values. Conversely, higher temperatures (\(\tau > 0.1\)) resulted in a significant increase in novel sequences, alongside greater divergence from the base sequences. These results suggest that temperature manipulation can strategically balance novelty and similarity, highlighting the effectiveness of isometric learning in structuring the latent space for protein design.

%% file: sections/6_conclusion.tex
We introduce \acf{PFM}, a novel framework for simulation-free training of generative models on data manifolds. By leveraging pullback geometry and isometric learning, \ac{PFM} allows for closed-form mappings on data manifolds while enabling precise interpolation and efficient generation. We demonstrated the effectiveness of \ac{PFM} through applications in synthetic protein dynamics and small protein generation, showcasing its potential in generating novel, property-specific samples through designable latent spaces. This approach holds significant promise for advancing generative modeling in fields like drug discovery and materials science, where precise and efficient sample generation is critical.

%% file: sections/A.1_background.tex
To achieve an interpolatable latent manifold we take a Riemannian geometric perspective. We start by introducing the notation and key concepts of differential and Riemannian geometry, for a formal description see \cite{lee2012smooth}. Second, we explain prior work on \acp{RAE} \cite{diepeveen2024pulling}, a framework for constructing interpolatable latent manifolds. Third, we summarize \ac{CFM} for generative modeling \cite{lipman2022flow}, a scalable way to train generative models in a simulation-free manner. Finally, we discuss how \ac{RFM} \cite{chen2024flow} generalize \ac{CFM} to Riemannian manifolds.

\subsection{Riemmanian Geometry}
\label{subsec:riemannian geometry}

A $d$-dimensional \textit{smooth manifold} $\mathcal{M}$ is a topological space that locally resembles $\R^d$, such that for each point $\boldsymbol{p} \in \mathcal{M}$, there exists a neighborhood $U$ of $\boldsymbol{p}$ and a \textit{homeomorphism} $\psi: U \to \R^d$, called a \textit{chart}. Then the \textit{tangent space} $ \mathcal{T}_{\boldsymbol{p}}\mathcal{M}$ at a point $\boldsymbol{p} \in \mathcal{M}$ is a vector space consisting of the tangent vectors at $\boldsymbol{p}$ representing the space of derivations at $\boldsymbol{p}$.

A \textit{Riemannian manifold} $\big(\mathcal{M}, (\cdot,\cdot)^{\mathcal{M}}\big)$ is a smooth manifold $\mathcal{M}$ equipped with a \textit{Riemannian metric} $(\cdot,\cdot)^{\mathcal{M}}$, which is a smoothly varying positive-definite inner product on the tangent space $\mathcal{T}_{\boldsymbol{p}}\mathcal{M} $ at each point $\boldsymbol{p}$. The Riemannian metric $(\cdot,\cdot)^{\mathcal{M}}$ defines the length of tangent vectors and the angle between them, thereby inducing a natural notion of distance on $\mathcal{M}$ based on the lengths of tangent vectors along curves between two points. 

The shortest path between two points on $\mathcal{M}$ is called a \textit{geodesic}, which generalizes the concept of straight lines in Euclidean space to curved manifolds. Geodesics on Riemannian manifold are found by minimizing
\begin{equation}
    E(\gamma) = \frac{1}{2} \int_0^1 \big(\dot{\gamma}(t),\dot{\gamma}(t)\big)_{\gamma(t)} dt,
\end{equation} 
whereas 
\begin{equation}
    L(\gamma) = \int_0^1 \sqrt{\big(\dot{\gamma}(t),\dot{\gamma}(t)\big)_{\gamma(t)}}dt
\end{equation} 
defines the distance between two points on the manifold. The \textit{exponential map}, 
\begin{equation}
    \exp_{\boldsymbol{p}}: \mathcal{T}_{\boldsymbol{p}}\mathcal{M} \to \mathcal{M},
\end{equation}
at $\boldsymbol{p}$ maps a tangent vector $\Xi_{\boldsymbol{p}} \in \mathcal{T}_{\boldsymbol{p}}\mathcal{M}$ to a point on $\mathcal{M}$ reached by traveling along the geodesic starting at $\boldsymbol{p}$ in the direction of $\Xi_{\boldsymbol{p}}$ for unit time. The \textit{logarithmic map}, 
\begin{equation}
    \log_{\boldsymbol{p}}: \mathcal{M} \to \mathcal{T}_{\boldsymbol{p}} \mathcal{M},
\end{equation} 
is the inverse of the exponential map, mapping a point $\boldsymbol{q} \in \mathcal{M}$ back to the tangent space $ \mathcal{T}_{\boldsymbol{p}}\mathcal{M}$ at $\boldsymbol{p}$. 

These names, 'exponential' and 'logarithmic' map, are geometric extensions of familiar calculus concepts. Just as the exponential function maps a number to a point on a curve, the exponential map on a manifold maps a direction and starting point to a location along a geodesic. Similarly, the logarithm in calculus reverses exponentiation, and the logarithmic map on a manifold reverses the exponential map, returning the original direction and distance needed to reach a specified point along the geodesic.

Assume $\big(\mathcal{M},(\cdot,\cdot)^{\mathcal{M}}\big)$ is a $d$-dimensional Riemannian manifold and a smooth diffeomorphism $\varphi: \R^d \rightarrow \mathcal{M}$, such that $\varphi(\R^d) \subseteq \mathcal{M}$ is geodesically convex, i.e., geodesics are uniquely defined on $\varphi(\R^d)$. We can then 
define the \textit{pullback metric} as
\begin{equation}
    (\Xi_{\mathbf{p}},\Phi_{\mathbf{p}})_{\mathbf{p}}^\varphi := \big(\varphi_\ast [\Xi_{\mathbf{p}}],\varphi_\ast [\Phi_{\mathbf{p}}] \big)_{\varphi(\mathbf{p})},
\end{equation}
for tangent vectors $\Xi_{\mathbf{p}}$ and $\Phi_{\mathbf{p}}$, where $\varphi_\ast$ is the pushforward. These mappings allow us to define all relevant geometric mappings in $\R^d$ in terms of manifold mappings on $\mathcal{M}$, see e.g. Proposition 2.1 of \cite{diepeveen2024pulling}:
\begin{enumerate}
    \item Distances $d^{\varphi}_{\R^d}: \R^d \times \R^d \rightarrow \R$ on $\big(\R^d, (\cdot,\cdot)^\varphi\big)$ are given by, 
    \begin{equation}
        d^{\varphi}_{\R^d}(\boldsymbol{x_i},\boldsymbol{x_j}) = d_{\mathcal{M}} \big( \varphi(\boldsymbol{x}_i),\varphi(\boldsymbol{x}_j) \big),
    \end{equation}
    \item Length-minimizing geodesics $\gamma^\varphi_{\boldsymbol{x}_i,\boldsymbol{x}_j} : [0,1] \rightarrow \R^d$ on $\big(\R^d,(\cdot,\cdot)^\varphi\big)$ are given by,
    \begin{equation}
        \gamma^\varphi_{\boldsymbol{x}_i,\boldsymbol{x}_j}(t) = \varphi^{-1}\big(\gamma^{\mathcal{M}}_{\varphi(\boldsymbol{x}_i),\varphi(\boldsymbol{x}_j)}(t)\big) \label{eq:geodesic mapping}
    \end{equation}
    \item Logarithmic maps $\log^\varphi_{\boldsymbol{x_i}}: \R^d \rightarrow \mathcal{T}_{\boldsymbol{x}_i}\R^d$ on $\big(\R^d,(\cdot,\cdot)^\varphi\big)$ are given by,
    \begin{equation}
        \log^\varphi_{\boldsymbol{x}_i}(\boldsymbol{x}_j)= \varphi_\ast^{-1}\left[\log^\mathcal{M}_{\varphi(\boldsymbol{x}_i)}\big(\varphi(\boldsymbol{x}_j)\big)\right]
    \end{equation}
    \item Exponential maps $\exp^\varphi_{\boldsymbol{x}_i}: \mathcal{G}_{\boldsymbol{x}_i} \rightarrow \R^d$ for $\mathcal{G}_{\boldsymbol{x}_i} := \log^\varphi_{\boldsymbol{x}_i}(\R^d) \subset \mathcal{T}_{\boldsymbol{x}_i}\R^d$ on $\big(\R^d,(\cdot,\cdot)^\varphi\big)$ are given by 
    \begin{equation}
        \exp^\varphi_{\boldsymbol{x}_i}(\Xi_{\boldsymbol{x}_i}) = \varphi^{-1}\big(\exp^\mathcal{M}_{\varphi(\boldsymbol{x}_i)}\left( \varphi_\ast [\Xi_{\boldsymbol{x}_i}]\right)\big)
    \end{equation}
\end{enumerate}

\begin{figure}[!h]
    \centering
    \includegraphics[width=0.9\linewidth]{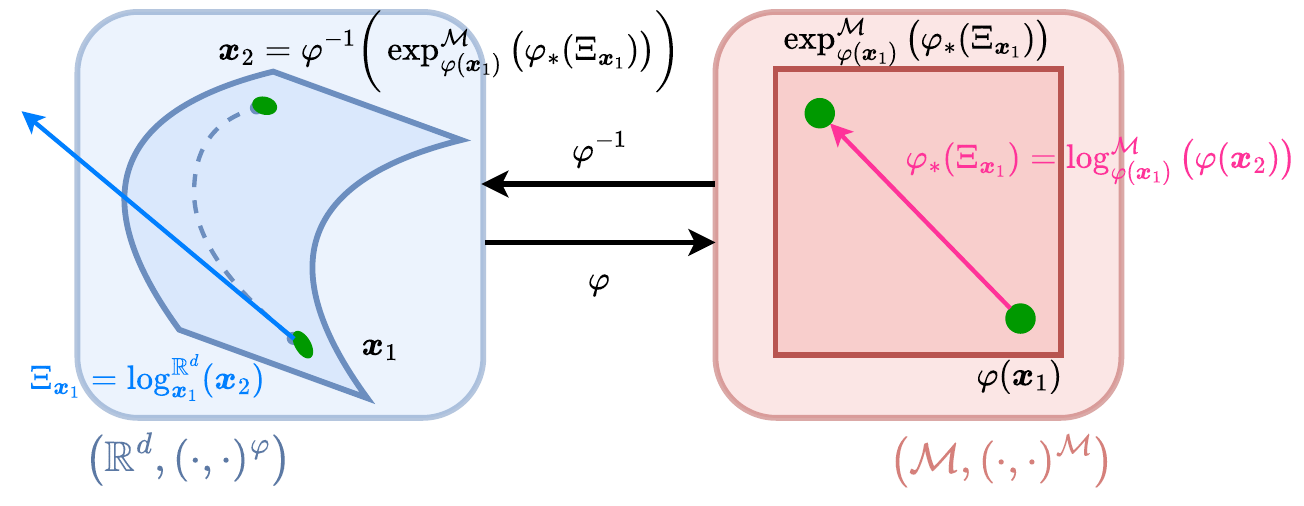}
    \caption{Example of pullback geometry for $\varphi: \R^d \rightarrow \mathcal{M}$ with $\mathcal{M}=\mathcal{M}_{d'} \times \R^{d-d'}$ for $\mathcal{M}_{d'}=\R^{d'}$, $d=3$ and $d'=2$. Samples $\varphi(\boldsymbol{x}_i)$ are close to elements of $\mathcal{M}_{d'}\times \boldsymbol{0}^{d-d'}$. }
    \label{fig:visual example pullback geometry}
\end{figure}

\noindent A visual example of pullback geometry is given in \autoref{fig:visual example pullback geometry}. Pullback geometry allows us to remetrize all of space $\R^d$, including the data manifold $\mathcal{D} \subset \R^d$, through the pullback metric. We can use it to define geometric mappings on $\big(\R^d,(\cdot,\cdot)^\varphi \big)$, including geodesics (see \autoref{eq:geodesic mapping}), through geometric mappings on the latent manifold $\mathcal{M}$. Next, we summarize work on Riemannian Auto-Encoders, that leverage pullback geometry to create an interpolatable latent manifold.

\subsection{Riemannian Auto-Encoder}
\label{subsec:riemannian auto-encoder}

The goal of \acp{RAE} is to create a interpolatable latent representation of the data. This is achieved through data-driven (pullback) Riemannian geometry, encoding the data onto a latent manifold with known geometry. The benefit of this, is that interpolation on the data manifold corresponds to interpolation on the latent manifold. Resulting in a more interpretable latent space compared to traditional auto-encoders.

Similar as in \cite{diepeveen2024pulling}, we define a \ac{RAE} as a Riemannian Encoder $\text{RE}: \R^d \rightarrow \R^r$ and Riemannian Decoder $\text{RD}: \R^r \rightarrow \R^d$,
\begin{align}
    \text{RAE}(\boldsymbol{x}) &:= (RD \circ RE)(\boldsymbol{x}) \quad s.t.,\\
    \text{RE}(\boldsymbol{x})_k &:= (\log^\varphi_{\boldsymbol{z}}(\boldsymbol{x}),\boldsymbol{v}_{\boldsymbol{z}}^k)^\varphi_{\boldsymbol{z}} \text{ for } k =1,\dots r, \\
    \text{RD}(\boldsymbol{a}) &:= \exp^\varphi_{\boldsymbol{z}} \left(\sum_{k=1}^r \boldsymbol{a}_k \boldsymbol{v}_{\boldsymbol{z}}^k\right)
\end{align}
where $\boldsymbol{z}$ denotes a base point and $(\cdot,\cdot)^\varphi_{\boldsymbol{z}}$ the pullback metric at $\boldsymbol{z}$. Furthermore, 
\begin{equation}
    \boldsymbol{v}_{\boldsymbol{z}}^k := \sum_{l=1}^d \boldsymbol{W}_{lk} \Phi^l_{\boldsymbol{z}},
\end{equation} 
represents the basis vectors of the latent space in the tangent space $T_{\boldsymbol{z}} \R^d$. Let $\Phi_{\boldsymbol{z}}^l \in T_{\boldsymbol{z}} \R^d$ be an orthonormal basis in the tangent space at $\boldsymbol{z}$ with respect to $(\cdot,\cdot)^\varphi_{\boldsymbol{z}}$ and define 
\begin{equation}
    \boldsymbol{X}_{i,l} = \big(\log_{\boldsymbol{z}}^\varphi(\boldsymbol{x}^i),\Phi^l_{\boldsymbol{z}}\big)_{\boldsymbol{z}}^\varphi \text{ for } i=1,\dots,n \text{ and } l=1,\dots,d.
\end{equation}
We can compute $\boldsymbol{W}$ through a Singular Value Decomposition (SVD) of $\boldsymbol{X}$. 
\begin{equation}
    \boldsymbol{X} = \boldsymbol{U\Sigma W}^T,
\end{equation}
where $U \in \R^{N \times R}$, $\Sigma = \text{diag}(\sigma_1, \dots, \sigma_R) \in \R^{R \times R}$ with $\sigma_1 \geq \dots \geq \sigma_R$, $W \in \R^{d \times R}$ and where $R := \text{rank}(\boldsymbol{X})$. The first $r$ columns of $\boldsymbol{W}$, corresponding to the largest singular values, are selected to form the matrix $\boldsymbol{W} \in \R^{d \times r}$. This parameter $r$ allows one to set the dimensionality of the latent representation of the \ac{RAE}, if $r=d$ then the \ac{RAE} reduces to $\text{RAE}(\boldsymbol{x}) = \exp^\varphi_{\boldsymbol{z}}\big(\log^\varphi_{\boldsymbol{z}}(\boldsymbol{x})\big)$.

To learn a \ac{RAE}, one needs to first construct a diffeomorphism and define an objective function. In \cite{diepeveen2024pulling} diffeomorphisms are constructed by,
\begin{equation}
    \varphi := [\psi^{-1},\boldsymbol{I}_{d-d'}] \circ \phi \circ \boldsymbol{O} \circ T_{\boldsymbol{z}}, \label{eq:diffeomorphism construction diepeveen}
\end{equation}
where $\psi : U \rightarrow \R^{d'}$ is a chart on a (geodesically convex) subset $U \subset \mathcal{M}^{d'}$ of a $d'$-dimensional Riemannian manifold $(\mathcal{M}^{d'}, (\cdot, \cdot)_{\mathcal{M}_d'})$, $\phi : \R^d \rightarrow \R^d$ is a real-valued diffeomorphism, $\boldsymbol{O} \in \mathbb{O}(d)$ is an orthogonal matrix, and $T_{\boldsymbol{z}} : \R^d \rightarrow \R^d$ is given by $T_{\boldsymbol{z}}(\boldsymbol{x}) = \boldsymbol{x} - \boldsymbol{z}$. The learnable diffeomorphism $\varphi := \varphi_{\boldsymbol{\theta}}$ is constructed through parameterizing $\phi:=\phi_{\boldsymbol{\theta}}$ by an invertible residual network \cite{behrmann2019invertible}. 

\subsection{Learning Isometries with Riemannian Auto-Encoders}
\label{subsec:learning isometries with riemannian auto-encoders}

After constructing the diffeomorphism and Riemannian Auto-Encoder, one can learn an isometry by find the parameters $\boldsymbol{\theta}$ of $\varphi_{\theta}$ in \cite{diepeveen2024pulling} through minimizing the objective,
\begin{align}
    \mathcal{L}(\boldsymbol{\theta}) &= \frac{1}{N(N-1)} \sum_{i,j=1, i \neq j}^{N(N-1)} \big( d_{\R^d}^{\varphi_{\boldsymbol{\theta}}} (\boldsymbol{x}_i, \boldsymbol{x}_j) - d_{i,j} \big)^2 \tag{\textcolor{kth-blue}{\textbf{global isometry loss}}} \label{eq:global isometry loss willems method}\\
    & + \alpha_{\text{sub}} \frac{1}{N} \sum_{i=1}^N \left\| \begin{bmatrix}
        \boldsymbol{I}_{d-d'} & \emptyset \\
        \emptyset & \boldsymbol{0}_{d'}
    \end{bmatrix} (\phi_{\boldsymbol{\theta}} \circ \boldsymbol{O} \circ T_{\boldsymbol{z}})(\boldsymbol{x}_i) \right\|_1 \tag{\textcolor{kth-blue}{\textbf{submanifold loss}}} \label{eq:submanifold loss willems method}\\
    & + \alpha_{\text{iso}} \frac{1}{N} \sum_{i=1}^N \left\| \left( \left( \boldsymbol{e}^j, \boldsymbol{e}^{j'} \right)^{\varphi_{\boldsymbol{\theta}}}_{\boldsymbol{x}_i}\right)_{j,j'=1}^d - \boldsymbol{I}_d \right\|_F^2 \tag{\textcolor{kth-blue}{\textbf{local isometry loss}}} \label{eq:local isometry loss willems method},
\end{align}
where $\| \cdot \|_F$ is the Frobenius norm and $ \left(\left( \boldsymbol{e}^j, \boldsymbol{e}^{j'} \right)^{\varphi_{\boldsymbol{\theta}}}_{\boldsymbol{x}_i}\right)_{j,j'=1}^d$ denotes a $d$-dimensional matrix just as $(\boldsymbol{A}_{ij})^d_{i,j=1}$ denotes a matrix.

First, the \ref{eq:global isometry loss willems method} takes global geometry into account, ensuring that the learned distances under the diffeomorphism $\varphi_{\boldsymbol{\theta}}$ approximate the true pairwise distances $d_{i,j}$ between data points. Second, the \ref{eq:submanifold loss willems method} enforces that the data manifold is mapped to  $\mathcal{M} = \mathcal{M}^{d'} \times \R^{d-d'}$, preserving the submanifold structure of the data in the latent space. Finally, the \ref{eq:local isometry loss willems method} enforces local isometry, ensuring that small-scale distances and local geometry are preserved under the transformation, which is critical for maintaining the intrinsic geometric properties of the data during dimensionality reduction. For further details on the implementation and theoretical considerations, see \cite{diepeveen2024pulling}.

\newpage

\subsection{Conditional Flow Matching}
\label{subsec:conditional flow matching}

To achieve the goal of accurate generative modeling on data manifolds through isometric learning, we first need to understand generative modeling on Euclidean spaces. We do this through summarizing \ac{CFM} \cite{lipman2022flow}, a commonly used and effective framework for learning \acp{CNF} for generative modeling for Euclidean data \cite{chen2018neural}. \Ac{CFM} is a method designed to map a simple base distribution to a target data distribution by learning a time-dependent vector field. The fundamental goal of \ac{FM} is to align a target probability path $p_t(\boldsymbol{x})$ with a vector field $u_t(\boldsymbol{x})$, which generates the desired distribution. The \ac{FM} objective is defined as follows:

\begin{equation}
    \mathcal{L}_{\text{FM}}(\boldsymbol{\eta}) = \mathbb{E}_{t,p_t(\boldsymbol{x})}\| v_t(\boldsymbol{x};\boldsymbol{\eta}) - u_t(\boldsymbol{x}) \|^2,
\end{equation}
where $\boldsymbol{\eta}$ represents the learnable parameters of the neural network that parameterizes the vector field $v_t(\boldsymbol{x};\boldsymbol{\eta})$, and $t \sim \mathcal{U}(0, 1)$ is uniformly sampled. However, a significant challenge in \ac{FM} is the intractability of constructing the exact path $p_t(\boldsymbol{x})$ and the corresponding vector field $u_t(\boldsymbol{x})$. 

To address this \cite{lipman2022flow} introduce \ac{CFM}, a more practical approach by constructing the probability path and vector fields in a conditional manner. The \ac{CFM} objective is then formulated by marginalizing over the data distribution $q(\boldsymbol{x}_1)$ and considering the conditional probability paths:

\begin{equation}
    \mathcal{L}_{\text{CFM}}(\boldsymbol{\eta}) = \mathbb{E}_{t, q(\boldsymbol{x}_1), p_t(\boldsymbol{x}|\boldsymbol{x}_1)} \| v_t(\boldsymbol{x};\boldsymbol{\eta}) - u_t(\boldsymbol{x}|\boldsymbol{x}_1) \|^2. \label{eq:general cfm objective}
\end{equation}
A key result, as established in Theorem~2 of \cite{chen2024flow}, is that the gradients of the \ac{CFM} objective with respect to the parameters $\boldsymbol{\eta}$ are identical to those of the original \ac{FM} objective, i.e.,

\begin{equation}
    \nabla_{\boldsymbol{\eta}} \mathcal{L}_{\text{FM}}(\boldsymbol{\eta}) = \nabla_{\boldsymbol{\eta}} \mathcal{L}_{\text{CFM}}(\boldsymbol{\eta}),
\end{equation}
ensuring that optimizing the \ac{CFM} objective yields the same result as the original \ac{FM} objective. This enables effective train of the neural network without needing direct access to the intractable marginal probability paths or vector fields.

Given a sample $\boldsymbol{x}_1$ from the data distribution $q(\boldsymbol{x}_1)$, we define a conditional probability path $p_t(\boldsymbol{x}|\boldsymbol{x}_1)$ \footnote{In this work, we use two types of indexing: $x_t$ to denote time indices and $x_i$ for different data points. It should be clear from the context which indexing is being used.}. This path starts at $t=0$ from a simple distribution, typically a standard Gaussian, and approaches a distribution concentrated around $\boldsymbol{x}_1$ as $t \rightarrow 1$:

\begin{equation}
    p_t(\boldsymbol{x}|\boldsymbol{x}_1) = \mathcal{N}(\boldsymbol{x} | \mu_t(\boldsymbol{x}_1), \sigma_t(\boldsymbol{x}_1)^2 \mathbf{I}),
\end{equation}
where $\mu_t(\boldsymbol{x}_1): [0,1] \times \R^d \rightarrow \R^d$ is the time-dependent mean, and we denote the time-dependent standard deviation as $\sigma_t(\boldsymbol{x}_1): [0,1] \times \R \rightarrow \R_{>0}$. For simplicity, we set $\mu_0(\boldsymbol{x}_1) = \boldsymbol{0}$ and $\sigma_0(\boldsymbol{x}_1) = 1$, ensuring that all conditional paths start from the same standard Gaussian distribution. At $t = 1$, the path converges to a distribution centered at $\boldsymbol{x}_1$ with a small standard deviation $\sigma_{\text{min}}$.

The corresponding conditional vector field $u_t(\boldsymbol{x}|\boldsymbol{x}_1)$ can be defined by considering the flow:

\begin{equation}
    \chi_t(\boldsymbol{x}) = \sigma_t(\boldsymbol{x}_1) \boldsymbol{x} + \mu_t(\boldsymbol{x}_1),
\end{equation}
which maps a sample from the standard Gaussian to a sample from $p_t(\boldsymbol{x}|\boldsymbol{x}_1)$. The vector field $u_t(\boldsymbol{x}|\boldsymbol{x}_1)$ that generates this flow, as proven by \cite{lipman2022flow} in Theorem 3, is given by:

\begin{equation}
    u_t(\boldsymbol{x}|\boldsymbol{x}_1) = \frac{\sigma_t'(\boldsymbol{x}_1)}{\sigma_t(\boldsymbol{x}_1)} \left(\boldsymbol{x} - \mu_t(\boldsymbol{x}_1)\right) + \mu_t'(\boldsymbol{x}_1),
\end{equation}
where the primes denote derivatives with respect to time to stay consistent with the original papers notation.

In this work, we choose to use the \ac{OT} formulation of \ac{CFM}. Here, the mean $\mu_t(\boldsymbol{x}_1)$ and standard deviation $\sigma_t(\boldsymbol{x}_1)$ are designed to change linearly in time, offering a straightforward interpolation between the base distribution and the target distribution. Specifically, the mean and standard deviation are defined as:
\begin{equation}
    \mu_t(\boldsymbol{x}_1) = t \boldsymbol{x}_1, \quad \sigma_t(\boldsymbol{x}_1) = 1 - (1 - \sigma_{\text{min}})t.
\end{equation}
This linear path results in a vector field $u_t(\boldsymbol{x}|\boldsymbol{x}_1)$ given by:

\begin{equation}
    u_t(\boldsymbol{x}|\boldsymbol{x}_1) = \frac{\boldsymbol{x}_1 - (1 - \sigma_{\text{min}})\boldsymbol{x}}{1 - (1 - \sigma_{\text{min}})t}.
\end{equation}
The corresponding conditional flow that generates this vector field is:

\begin{equation}
    \chi_t(\boldsymbol{x}) = (1 - (1 - \sigma_{\text{min}})t)\boldsymbol{x} + t\boldsymbol{x}_1.
\end{equation}
This \ac{OT} path is optimal in the sense that it represents the displacement map between the two Gaussian distributions $p_0(\boldsymbol{x}|\boldsymbol{x}_1)$ and $p_1(\boldsymbol{x}|\boldsymbol{x}_1)$ \cite{lipman2022flow}.

The final \ac{CFM} loss under this \ac{OT} formulation is derived by substituting the above vector field and flow into the general \ac{CFM} objective (\autoref{eq:general cfm objective}) and reparameterizing $p_t(\boldsymbol{x}|\boldsymbol{x}_1)$ in terms of $\boldsymbol{x}_0$. This yields the following objective function:

\begin{equation}
    \mathcal{L}_{\text{CFM}}(\boldsymbol{\eta}) = \mathbb{E}_{t, q(\boldsymbol{x}_1), p(\boldsymbol{x}_0)}\left\| v_t(\chi_t(\boldsymbol{x}_0); \boldsymbol{\eta}) - \frac{\boldsymbol{x}_1 - (1 - \sigma_{\text{min}})\boldsymbol{x}_0}{1 - (1 - \sigma_{\text{min}})t} \right\|^2.
\end{equation}
This formulation is advantageous because the \ac{OT} paths ensure that particles move in straight lines and with constant speed, leading to simpler and more efficient regression tasks compared to traditional diffusion-based methods. We use the \acs{OT}-\acs{CFM} objective in this work when we refer to \ac{CFM}. 

\newpage

\subsection{Riemannian Flow Matching}
\label{subsec:riemannian conditional flow matching}

The next step toward generation on data manifolds is understanding generation on manifolds with closed form geometric mappings. \Ac{RFM} aims to do exactly this by generalizing \ac{CFM} to Riemannian manifolds \cite{chen2024flow}. Assume a complete, connected and smooth manifold $\mathcal{M}$ endowed with a Riemannian metric $(\cdot,\cdot)^{\mathcal{M}}$. We are given a set of training samples $\boldsymbol{x}_1 \in \mathcal{M}$ from some unknown data distribution $q(\boldsymbol{x}_1)$ on the manifold. Then the goal is to learn a parametric map $\rho: \mathcal{M} \rightarrow \mathcal{M}$ that pushes a simple base distribution $p$ to the data distribution $q$. To achieve RFM \cite{chen2024flow} reparameterize the conditional flow as
\begin{equation}
    \boldsymbol{x}_t = \chi_t(\boldsymbol{x}_0|\boldsymbol{x}_1),
\end{equation}
where $\chi_t(\boldsymbol{x}_0|\boldsymbol{x}_1)$ is the solution to the \ac{ODE} defined by a time-dependent conditional vector field $u_t(\boldsymbol{x}|\boldsymbol{x}_1) \in \mathcal{T}_{\boldsymbol{x}}\mathcal{M}$ that is tangent to the manifold $\mathcal{M}$. The initial condition is set as $\chi_0(\boldsymbol{x}_0|\boldsymbol{x}_1) = \boldsymbol{x}_0$. 

This formulation leads to the RFM objective, which ensures that the vector field $u_t(\boldsymbol{x}|\boldsymbol{x}_1)$ learned by the model lies entirely within the tangent space of the manifold at each point $\boldsymbol{x}_t \in \mathcal{M}$:
\begin{equation}
    \mathcal{L}_{\text{RFM}}(\boldsymbol{\eta}) = \mathbb{E}_{t, q(\boldsymbol{x}_1), p(\boldsymbol{x}_0)} \Bigl(\| v_t(\boldsymbol{x}_t; \boldsymbol{\eta}) - u_t(\boldsymbol{x}_t|\boldsymbol{x}_1) \|_{\mathbf{x}_t}^{\mathcal{M}}\Bigr)^2,
\end{equation}
where $\|\cdot \|_{\mathcal{M}}$ is the norm enduced by the Riemannian metric $(\cdot,\cdot)^{\mathcal{M}}$. For manifolds with closed-form geodesic expressions, a simulation-free objective can be formulated using exponential and logarithmic maps. This approach allows models to be trained without numerically simulating particle trajectories, leveraging closed-form geodesics and mappings to directly compute vector fields and transport paths. In this case, $\boldsymbol{x}_t$ can be defined by the geodesic between $\boldsymbol{x}_1$ and $\boldsymbol{x}_0$ and can be explicitly expressed as
\begin{equation}
    \boldsymbol{x}_t = \gamma^{\mathcal{M}}_{\boldsymbol{x}_1,\boldsymbol{x}_0}\big(\kappa(t)\big) = \exp^{\mathcal{M}}_{\boldsymbol{x}_1} \big(\kappa(t)\log^{\mathcal{M}}_{\boldsymbol{x}_1}(\boldsymbol{x_0})\big),
\end{equation}
with monotonically decreasing differentiable function $\kappa(t)$ satisfying $\kappa(0) = 1$ and $\kappa(1) = 0$ acting as a scheduler. Furthermore, the tangent vector field $u_t(\boldsymbol{x}|\boldsymbol{x}_1)$ can be evaluated through,
\begin{equation}
    u_t(\boldsymbol{x}|\boldsymbol{x}_1) = \dot{\gamma}^{\mathcal{M}}_{\boldsymbol{x}_1,\boldsymbol{x}_0}\big(\kappa(t)\big) = \frac{d}{dt} \exp^{\mathcal{M}}_{\boldsymbol{x}_1} \big(\kappa(t)\log^{\mathcal{M}}_{\boldsymbol{x}_1}(\boldsymbol{x_0})\big)
\end{equation}
The objective function is then given by:
\begin{equation}
    \mathcal{L}_{\text{RFM}}(\boldsymbol{\eta}) = \mathbb{E}_{t, q(\boldsymbol{x}_1), p(\boldsymbol{x}_0)} \Bigl(\left\| v_t(\gamma^{\mathcal{M}}_{\boldsymbol{x}_0,\boldsymbol{x}_1}\big(\kappa(t)\big); \boldsymbol{\eta}) - \dot{\gamma}^{\mathcal{M}}_{\boldsymbol{x}_0,\boldsymbol{x}_1}\big(\kappa(t)\big)\right\|_{\mathbf{p}}^{\mathcal{M}}\Bigr)^2
    \label{eq:riemannian conditional flow mathcing objective}
\end{equation}
Constructing a simulation-free objective for RFM on general geometries presents significant challenges due to the absence of closed-form expressions for essential geometric operations, such as exponential and logarithmic maps, or geodesics. These operations are crucial for defining and efficiently evaluating the objective but are often computationally intensive to approximate without closed-form solutions. For a list of examples of manifolds with closed-form geometric mappings, see the appendix of \cite{chen2024flow}.

In the absence of such closed-form solutions, existing methods tackle these difficulties by either learning a metric that constrains the generative trajectory to align with the data support \cite{kapusniak2024metric} or by assuming a metric with easily computable geodesics on the data manifold \cite{chen2024flow}. However, learning a metric can be problematic as it may lead to overfitting or fail to capture the true geometry of the data, particularly when the data manifold is complex or poorly understood. On the other hand, assuming a simple metric with computable geodesics can oversimplify the problem, resulting in models that inadequately represent the underlying data structure. To overcome these challenges, we introduce Pullback Flow Matching in \autoref{sec:pullback flow matching}. 

%% file: sections/A.2_closed_form_manifold_mappings.tex
In \autoref{tab:manifold mappings with closed form expressions}, we include a table of manifolds and their closed-form geometric mappings from \cite{chen2024flow}. These mappings provide explicit expressions for the exponential and logarithmic maps, which construct local coordinate charts. Given a reference point $\boldsymbol{x} \in \mathcal{M}$, the chart $\psi$ is defined as $\psi(\boldsymbol{y}) = \log_{\boldsymbol{x}}^{\mathcal{M}}(\boldsymbol{y})$, mapping points on the manifold to the tangent space at $\boldsymbol{x}$. The inverse chart is the exponential map, $\psi^{-1}(\Xi) = \exp_{\boldsymbol{x}}^{\mathcal{M}}(\Xi)$, ensuring a local diffeomorphism.

\begin{table}[h!]
    \centering
    \caption{Riemannian manifolds with closed-form geometric mappings. The table includes exponential maps, logarithm maps, and inner products, the operator $\oplus$ denotes Möbius addition. For any points $\boldsymbol{x}, \boldsymbol{y} \in \mathcal{M}$ and tangent vectors $\Xi, \Phi \in T_{\boldsymbol{x}} \mathcal{M}$, we use the notation $\| \cdot \|^2_{\mathcal{M}} = (\cdot, \cdot )^{\mathcal{M}}$ for the Riemannian metric. This table is adapted from \cite{chen2024flow}.}
    \label{tab:manifold mappings with closed form expressions}
    \resizebox{\textwidth}{!}{
        \begin{tabular}{lccc}
            \toprule
            \textbf{Manifold} $\mathcal{M}$ & $\exp^{\mathcal{M}}_{\boldsymbol{x}}(\Xi_{\boldsymbol{x}})$ & $\log^{\mathcal{M}}_{\boldsymbol{x}}(\boldsymbol{y})$ & $(\Xi, \Phi)^{\mathcal{M}}$ \\ 
            \midrule
            $N$-D sphere $\{\boldsymbol{x} \in \mathbb{R}^{N+1} : \|\boldsymbol{x}\|_2 = 1\}$ & $x \cos (\|\Xi_{\boldsymbol{x}}\|_2) + \frac{\Xi_{\boldsymbol{x}}}{\|\Xi_{\boldsymbol{x}}\|_2} \sin (\|\Xi_{\boldsymbol{x}}\|_2)$ & $\arccos\big(( \boldsymbol{x}, \boldsymbol{y})\big) \frac{P_{\boldsymbol{x}} (\boldsymbol{y} - \boldsymbol{x})}{\|P_{\boldsymbol{x}} (\boldsymbol{y} - \boldsymbol{x})\|_2}$ & $(\Xi_{\boldsymbol{x}}, \Phi_{\boldsymbol{x}})^{\mathcal{M}}$ \\ 
            $N$-D flat tori $[0, 2\pi]^N$ & $(\boldsymbol{x} + \Xi_{\boldsymbol{x}}) \; \% \; (2\pi)$ & $\arctan2 \big(\sin(\boldsymbol{y} - \boldsymbol{x}), \cos(\boldsymbol{y} - \boldsymbol{x})\big)$ & $(\Xi_{\boldsymbol{x}},\Phi_{\boldsymbol{x}})^{\mathcal{M}}$ \\ 
            $N$-D Hyperbolic $\{\boldsymbol{x} \in \mathbb{R}^N : \|\boldsymbol{x}\|_2 < 1\}$ & $ \boldsymbol{x} \oplus \left( \tanh \left(\frac{\|\Xi_{\boldsymbol{x}}\|_2}{1 - \|\boldsymbol{x}\|_2^2}\right) \frac{\Xi_{\boldsymbol{x}}}{\|\Xi_{\boldsymbol{x}}\|_2} \right)$ & $\left(1 - \|\boldsymbol{x}\|_2^2\right) \tanh^{-1}\left(\|{-\boldsymbol{x} \oplus \boldsymbol{y}}\|_2\right) \frac{-\boldsymbol{x} \oplus \boldsymbol{y}}{\|{-\boldsymbol{x} \oplus \boldsymbol{y}}\|_2}$ & $\frac{4}{(1 - \|\boldsymbol{x}\|_2^2)^2} (\Xi_{\boldsymbol{x}}, \Phi_{\boldsymbol{x}})^{\mathcal{M}}$ \\ 
            $N \times N$ SPD matrices & $\boldsymbol{X}^{\frac{1}{2}} \exp\{\boldsymbol{X}^{-\frac{1}{2}} \boldsymbol{U} \boldsymbol{X}^{-\frac{1}{2}}\} \boldsymbol{X}^{\frac{1}{2}}$ & $\boldsymbol{X}^{\frac{1}{2}} \log\{\boldsymbol{X}^{-\frac{1}{2}} \boldsymbol{Y} \boldsymbol{X}^{-\frac{1}{2}}\} \boldsymbol{X}^{\frac{1}{2}}$ & $\text{tr}\left( \boldsymbol{X}^{-1} \Xi_{\boldsymbol{X}} \boldsymbol{X}^{-1} \Phi_{\boldsymbol{X}} \right)$ \\ 
            \bottomrule
        \end{tabular}
    }
\end{table}

%% file: sections/A.3_proof_that_neural_odes_parameterize_a_diffeomorphisms.tex
We can verify that this defines a diffeomorphism by using Theorem C.15 of \cite{younes2010shapes}. According to Theorem~C.15, for $\phi_{\boldsymbol{\theta}}$ to be a diffeomorphism, the vector field $f$ must satisfy $f \in L^1([0,1], C^1_{(0)}(\Omega, \mathbb{B}))$, where $\Omega$ is the domain of the vector field and $\mathbb{B}$ is a Banach space representing the target space.

In our case, $f$ is composed of smooth and continuously differentiable functions due to the \ac{MLP} parameterization, ensuring $f$ is also smooth and continuously differentiable. Additionally, we enforce local isometry by regularizing the Jacobian of $f_{\boldsymbol{\theta}}$, which guarantees local regularity of $f$ in the data domain (see \ref{eq:stability regularization}). Thus, $f$ meets the required conditions and $\phi_{\boldsymbol{\theta}}$ defines a proper diffeomorphism.

%% file: sections/A.5_manifold_and_metric_selection.tex
Isometric learning requires three key choices to be made, first one needs to choose the Riemannian metric of the data manifold $(\cdot,\cdot)^{\mathcal{D}}$, second one needs to choose both the latent (sub)manifold and its Riemannian metric $\big(\mathcal{M}_{d'},(\cdot,\cdot)^{\mathcal{M}_{d'}}\big)$ and finally one needs to choose the dimensionality $d'$. Technically one also needs to assume a metric on $\R^{d-d'}$, but in this work we assume a Euclidean metric $(\cdot,\cdot)_2$ throughout all our experiments.

There are several options when selecting the metric on the data manifold $(\cdot,\cdot)^{\mathcal{D}}$. One can choose for example a locally euclidean approximation through Isomap \cite{tenenbaum2000global} or a more noise-robust geodesic approximation \cite{little2022balancing}. One can also design a metric to create a latent space \footnote{In this text we refer to the latent space as the concept in machine and representation learning, technically its a latent manifold endowed with a Riemannian metric, not a vector space.} structured based on properties of the data one cares about, we show how in \autoref{subsec:designable latent spaces for protein design}. In this work, we focus on using a proper metric and defer the exploration of learning with pseudo-metrics to future research.

When selecting a latent Riemannian (sub)manifold and metric it is crucial to select $\mathcal{M}_{d'}$ such that $\mathcal{M}= \mathcal{M}_{d'} \times \R^{d-d'}$ it is diffeomorphic to the data manifold $\mathcal{D}$. This ensures that the latent space of the RAE can effectively capture the intrinsic structure of the data. The manifold should be chosen based on its ability to accommodate the data’s periodicity, curvature, and dimensionality. This alignment is essential for accurately representing the data manifold within the latent space. Unless otherwise stated we assume $\mathcal{M}_{d'} = \R^{d'}$. Additionally, one should select the Riemannian geometry of $\big(\mathcal{M}_{d'},(\cdot,\cdot)^{\mathcal{M}_{d'}}\big)$ such that geometric mappings can be explicitly defined in closed form. A list of manifolds with closed form geometric mappings can be found in the appendix of \cite{chen2024flow}. Unless otherwise states we select $(\cdot,\cdot)^{\mathcal{M}_{d'}}=(\cdot,\cdot)_2$.

Finally, $d'$, the dimensionality of the latent space, is a hyperparameter that could be tuned through iterative testing. Techniques such as Isomap \cite{tenenbaum2000global} or equivalents on other manifolds such as hyperbolic space \cite{cvetkovski2011multidimensional} can be employed to evaluate various dimensional and Riemannian geometric settings and determine the optimal $d'$ that balances model complexity with the ability to accurately capture the data manifold’s structure.

%% file: sections/A.4_data_description.tex
In this work we use several datasets, synthetic, simulated and experimental. Here we describe them in order of appearance in the experiments. 

\subsection{ARCH Dataset}
We create a dataset in the spirit of \cite{tong2020trajectorynet}. We sample $n=500$ data points uniform on the line $[-1,1]$ ($x_i \sim \mathcal{U}(-1,1)$), wrap this line around the unit half circle and add normally distributed noise with $\sigma=0.1$, i.e. \begin{equation}
    y_{i,1} = \sin (0.5 \pi x_i) + a_{i,1}, \quad y_{i,2} = \cos (0.5 \pi x_i) + a_{i,2} \text{ for } a_{i,j} \sim \mathcal{N}(0,0.1^2).
\end{equation}
An example of the dataset can be found in \autoref{fig:geodesic example figure}.

\subsection{\Acf{AK}}
\begin{figure}[h!]
    \centering
    \begin{minipage}{0.18\linewidth}
        \centering
        \includegraphics[width=\linewidth]{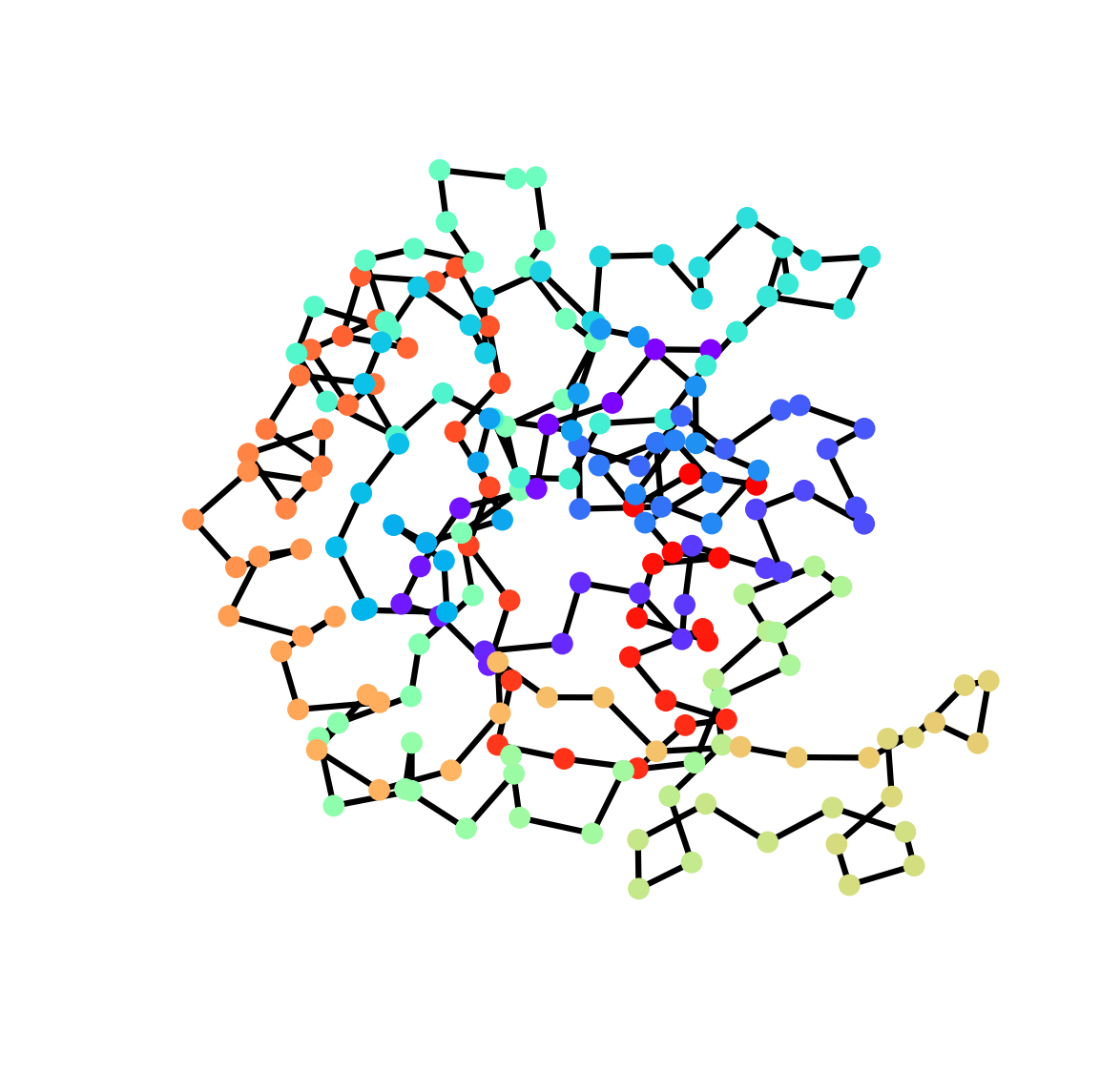}
        \caption*{$t=0$}  
    \end{minipage}%
    \hfill
    \begin{minipage}{0.18\linewidth}
        \centering
        \includegraphics[width=\linewidth]{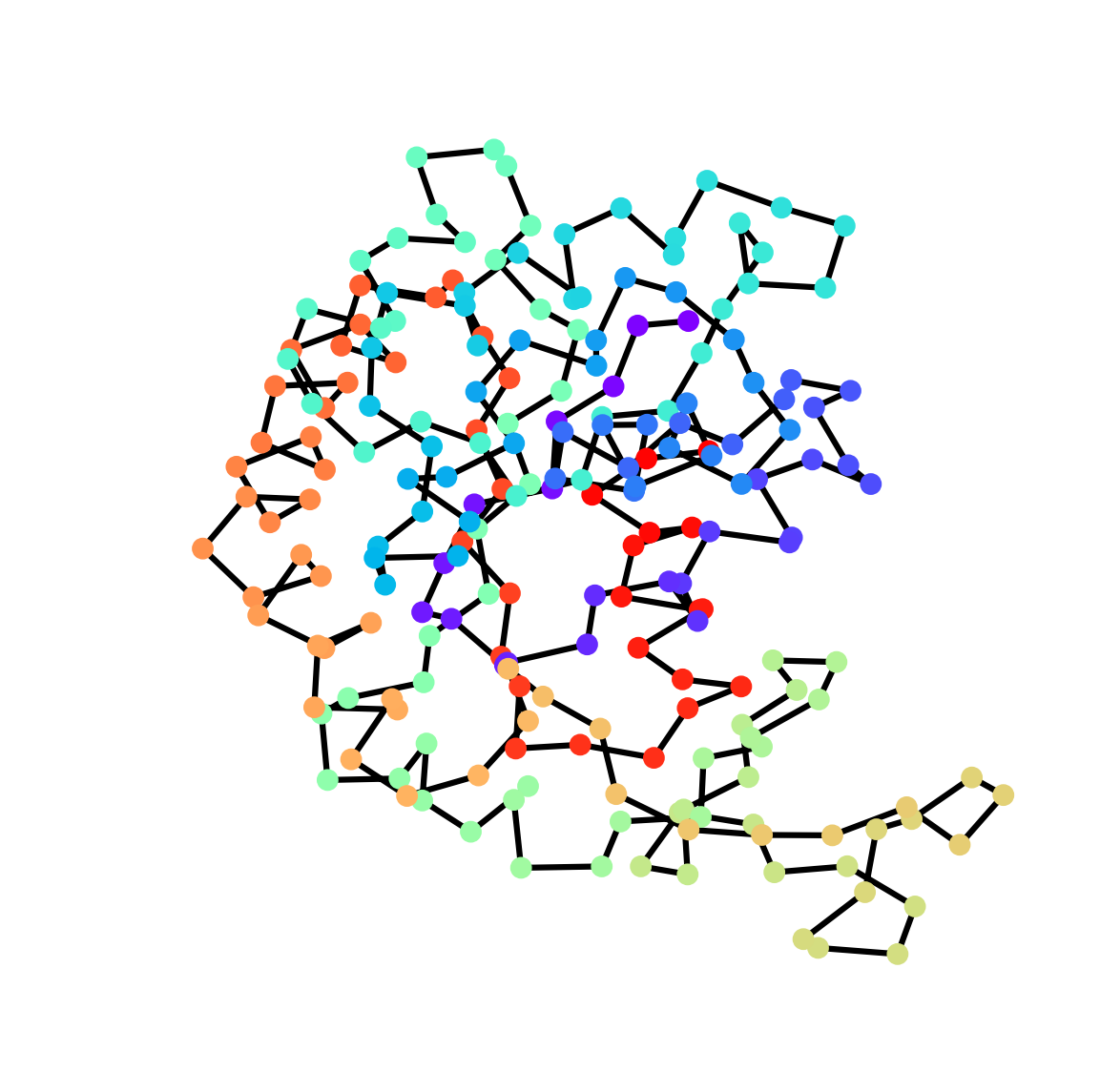}
        \caption*{$t=0.25$}  
    \end{minipage}%
    \hfill
    \begin{minipage}{0.18\linewidth}
        \centering
        \includegraphics[width=\linewidth]{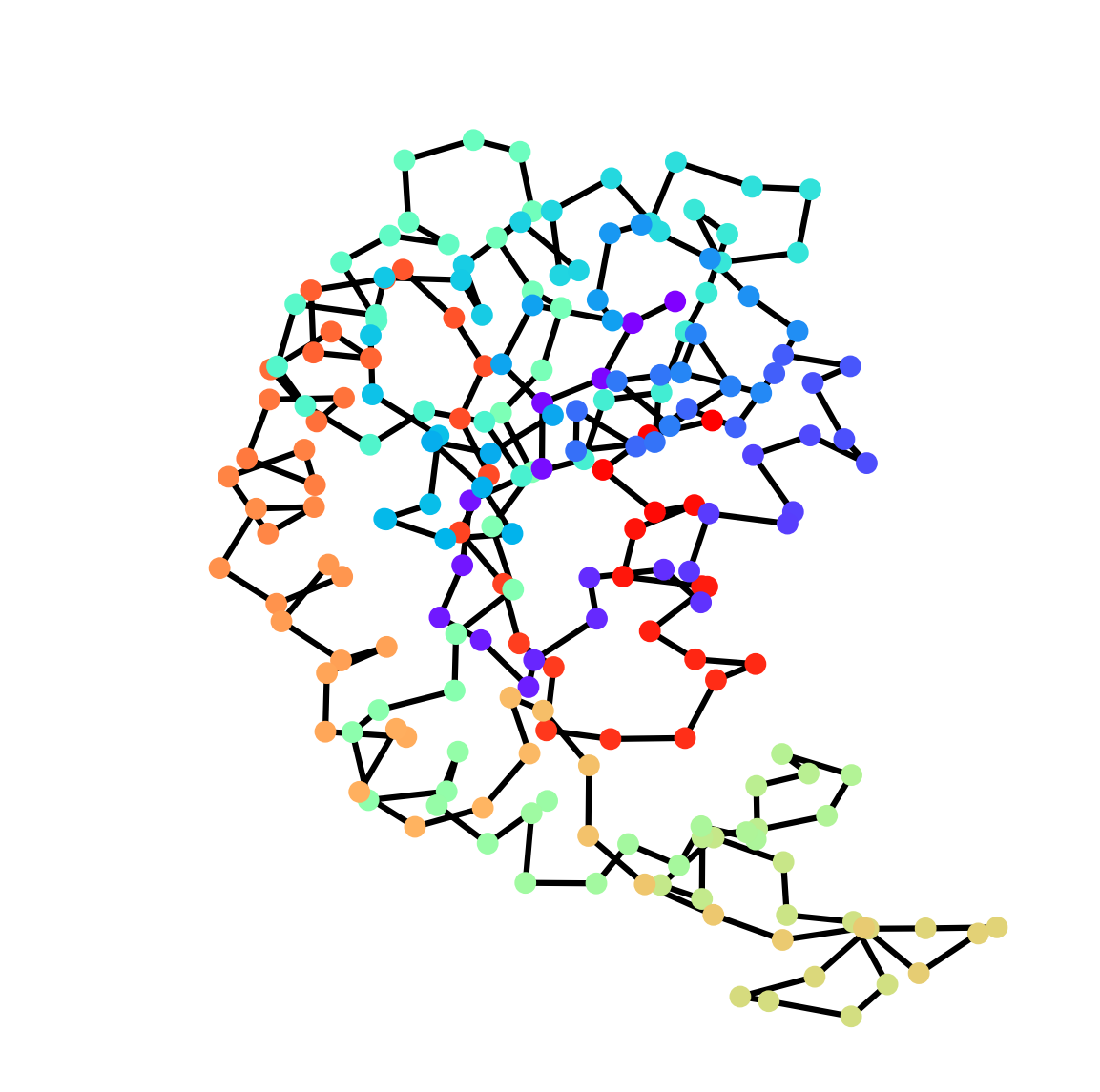}
        \caption*{$t=0.5$}  
    \end{minipage}%
    \hfill
    \begin{minipage}{0.18\linewidth}
        \centering
        \includegraphics[width=\linewidth]{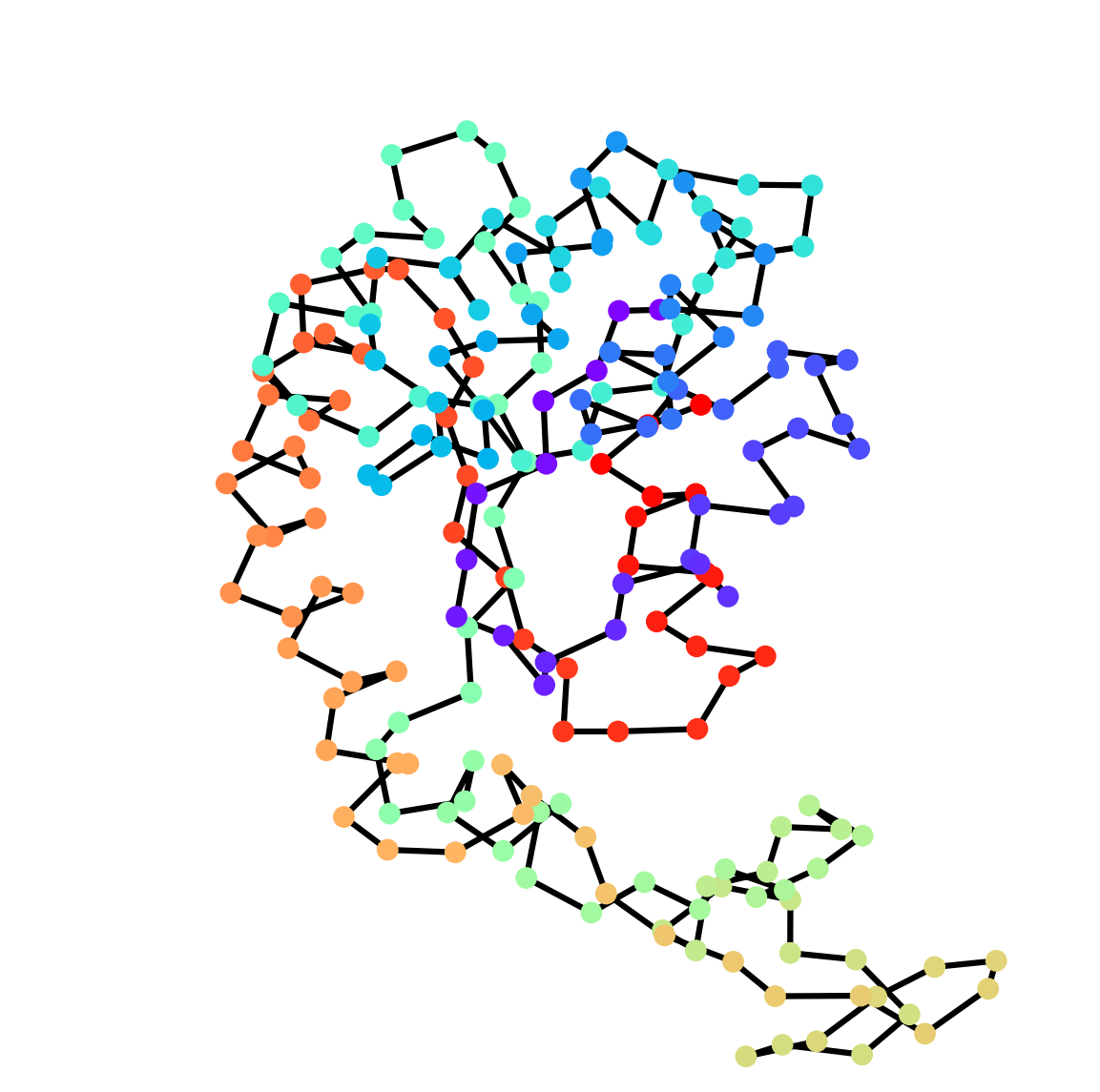}
        \caption*{$t=0.75$}  
    \end{minipage}%
    \hfill
    \begin{minipage}{0.18\linewidth}
        \centering
        \includegraphics[width=\linewidth]{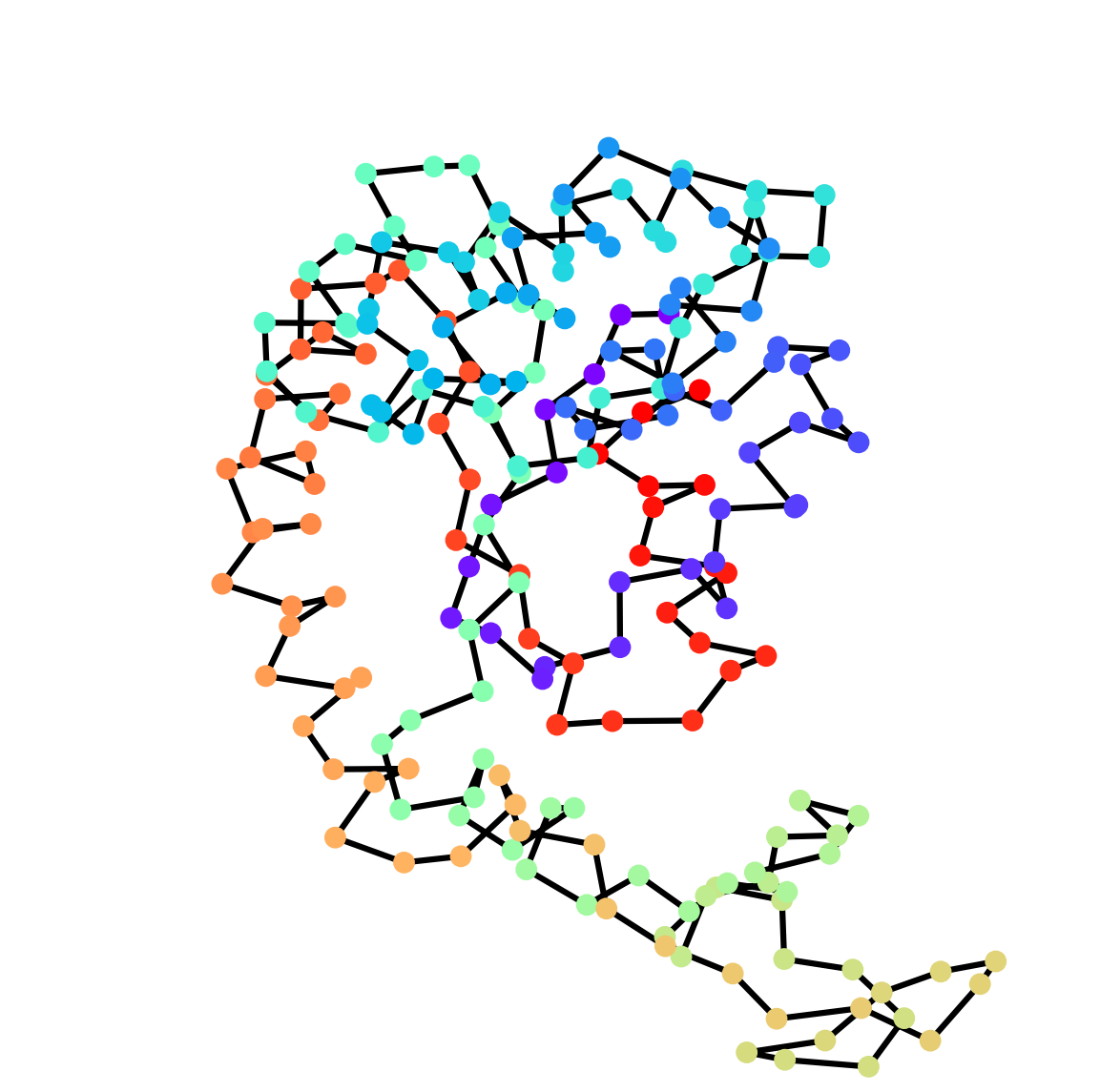}
        \caption*{$t=1$}  
    \end{minipage}
    \caption{Example of the open-to-close transition of adynalate kinase protein.}
    \label{fig:example_AK_figure}
\end{figure}

We consider the time-normalized open-to-close transition of \ac{AK}. This is a dataset from coarse-grained molecular dynamics simulations consisting of $n=102$ conformations of $214$ amino-acids in $3D$, samples of the trajectory can be found in \autoref{fig:example_AK_figure}.

\subsection{Intestinal Fatty Acid Binding Protein}
The second protein dynamics dataset is that of $n=500$ conformations of \ac{I-FABP} in water. The datasets comes from simulations in CHARMM of $500$ picoseconds (ps) with a $2$ femtoseconds (fs) timestep. The data can be found on \href{https://www.mdanalysis.org/MDAnalysisData/ifabp_water.html}{mdanalysis.org}.

\subsection{\Acs{GRAMPA} Dataset}
The \acf{GRAMPA} dataset \cite{witten2019deep} is a compilation of peptides and their antimicrobial activity against various bacteria, including \textit{E.~coli} and \textit{P.~aeruginosa}. It includes data on peptide sequences, target bacteria, bacterial strains, and \ac{MIC} values, with additional columns providing details on sequence modifications and data sources. The dataset was created to support deep learning models aimed at predicting the antimicrobial effectiveness of peptides. The dataset is available \href{https://github.com/zswitten/Antimicrobial-Peptides}{here}. In our experiments we follow the preprocessing pipeline from \cite{szymczak2023discovering} and use only the sequence data and the corresponding \ac{MIC} scores. After preprocessing we are left with $n=3444$ sequences of maximum $25$ amino-acids with tested antimicrobial activity against E. coli.

%% file: sections/A.6_training_procedure.tex
We explain the training procedure and hyperparameter settings for each of the experiments in \autoref{sec:experiments} in further detail for reproducibility. In all experiments the datasets where split into train and test sets. We apply early stopping and present the model with the lowest average loss on the test data. 

\subsection{Ablation Study}

For details of hyperparameter settings for the ablation study see \autoref{tab:ablation study hyperparameters}. 

\begin{table}[h!]
    \centering
    \small
    \caption{Hyperparameter settings for ablation study of \ac{RAE} on the \ac{ARCH}, \ac{AK} and \ac{I-FABP} datasets.}
    \label{tab:ablation study hyperparameters}
    \begin{tabular}{lccc}
        \toprule
        \textbf{Hyperparameter} & \textbf{\acs{ARCH}} & \textbf{\acs{AK}} & \textbf{\acs{I-FABP}}\\ 
        \midrule
        Epochs & $1000$ & $1000$ & $1000$\\ 
        Learning Rate & $0.0001$ & $0.0001$ & $0.0001$ \\ 
        Optimizer & \texttt{Adam} & \texttt{Adam} & \texttt{Adam} \\
        Train/Test Split & $0.8/0.2$ & $0.8/0.2$ & $0.8/0.2$ \\ 
        $n_{steps}$ & $10$ & $10$ & $10$ \\ 
        Seed & $0$ & $0$ & $0$\\ 
        Number of Layers & $5$ & $5$ & $5$ \\ 
        $\alpha_1$ & $1.0$ & $1.0$ & $1.0$ \\ 
        $\alpha_2$ & $[0.0,1.0]$ & $[0.0,5.0]$ & $[0.0,5.0]$ \\ 
        $\alpha_3$ & $1.0$ & $1.0$ & $1.0$ \\ 
        $\alpha_4$ & $[0.0,0.01]$ & $[0.0,0.005]$ & $[0.0,0.1]$ \\ 
        $d'$ & $1$ & $1$ & $1$\\ 
        Hidden Units & $64$ & $214\cdot 3+1$ & $131\cdot3+1$ \\ 
        Number of Neighbors & $5$ & $2$ & $4$\\ 
        Batch Size & $64$ & $16$ & $64$ \\ 
        Warmup & $50$ & $400$ & $200$ \\ 
        \bottomrule
    \end{tabular}
\end{table}
\noindent Specific hyperparameters worth mentioning are $n_{steps}$ which is the number of Runge-Kutta steps we use in our \ac{NODE}, \textit{Number of Layers} is the number of layers of the \ac{MLP} with swish activation function for the vector field of the \ac{NODE}. The \textit{Number of Neighbors} is the hyperparameter used to calculate the shortest paths over the nearest neighbors graph for the Isomap geodesics in \texttt{sklearn} and the \textit{Warmup} is the number of epochs we train with $\alpha_1,\alpha_2=0$ to first learn a lower dimensional representation.

\newpage 

\subsection{Interpolation Experiments}

For details of hyperparameter settings for the interpolation experiments of $(\cdot,\cdot)^{\mathcal{M}}$- and $(\cdot,\cdot)^{\mathcal{M}_{d'}}$-interpolation see \autoref{tab:interpolation experiments hyperparameters rae} and for the ($\beta$-)\acsp{VAE} see \autoref{tab:interpolation experiments hyperparameters (beta-)vae}.

\begin{table}[h!]
    \centering
    \small
    \caption{Hyperparameter settings for interpolation experiments for $(\cdot,\cdot)^{\mathcal{M}}$- and $(\cdot,\cdot)^{\mathcal{M}_{d'}}$-interpolation on the \ac{ARCH}, \ac{AK} and \ac{I-FABP} datasets.}
    \label{tab:interpolation experiments hyperparameters rae}
    \begin{tabular}{lcccc}
        \toprule
        \textbf{Hyperparameters} & \textbf{\acs{ARCH}} & \textbf{\acs{AK}} & \textbf{\acs{I-FABP}}\\
        \midrule
        Epochs & $1000$ & $1000 $ & $1000$\\ 
        Learning Rate & $0.0001$ & $0.0001$ & $0.0001$\\
        Optimizer & \texttt{Adam} & \texttt{Adam} & \texttt{Adam}\\
        Train/Test Split & $0.8/0.2$ & $0.8/0.2$ & $0.8/0.2$\\ 
        $n_{steps}$ & $10$ & $10$ & $10$\\ 
        Seed & $0$ & $0$  & $0$\\ 
        Number of Layers & $5$ & $5$ & $5$\\ 
        $\alpha_1$ & $1.0$ & $1.0$ & $1.0$ \\ 
        $\alpha_2$ & $5.0$ & $5.0$ & $5.0$ \\ 
        $\alpha_3$ & $1.0$ & $1.0$ & $1.0$ \\ 
        $\alpha_4$ & $0.001$ & $0.005$ & $0.1$ \\ 
        $d'$ & $1$ & $1$ & $1$ & $5$\\ 
        Hidden Units & $64$ & $214\cdot 3+1$  & $131 \cdot 3+1$\\ 
        Number of Neighbors & $5$ & $2$ & $4$\\ 
        Batch Size & $64$ & $16$  & $64$\\ 
        Warmup & $50$ & $400$ & $200$\\ 
        $n_{parameters}$ & $17282$ & $2486480$ & $934961$\\
        \bottomrule
    \end{tabular}
\end{table}

\begin{table}[h!]
        \centering
        \small
        \caption{Hyperparameter settings for interpolation experiments for ($\beta$-)\acs{VAE} on the \ac{ARCH} dataset. \Acp{VAE} have $\beta=1.0$, $\beta$-\acsp{VAE} have $\beta=10.0$.}
        \label{tab:interpolation experiments hyperparameters (beta-)vae}
        \begin{tabular}{lcccc}
            \toprule
            \textbf{Hyperparameters} & \textbf{\acs{ARCH}} & \textbf{\acs{AK}} & \textbf{\acs{I-FABP}}\\
            \midrule
            Epochs & $1000$ & $1000$ & $1000$\\ 
            Learning Rate & $0.0001$ & $0.0001$ & $0.0001$ \\ 
            Optimizer & \texttt{Adam} & \texttt{Adam} & \texttt{Adam} \\
            Train/Test Split & $0.8/0.2$ & $0.8/0.2$ & $0.8/0.2$\\ 
            Seed & $0$ & $0$ & $0$ \\ 
            Number of Encoder Layers & $5$ & $5$ & $5$\\ 
            Number of Decoder Layers & $5$ & $5$ & $5$\\ 
            Hidden Units & $64$ & $214\cdot3+1$ & $131\cdot3+1$\\ 
            Beta & $[1.0,10.0]$ & $[1.0,10.0]$ & $[1.0,10.0]$\\ 
            $d'$ & $1$ & $1$ & $1$\\ 
            Batch Size & $64$ & $16$ & $64$ \\ 
            $n_{parameters}$ & $34184$ & $4555655$ & $1712324$ \\
            \bottomrule
        \end{tabular}
\end{table}

\newpage

\subsection{Generation Experiments}

\begin{table}[h!]
    \centering
    \small
    \caption{Hyperparameter settings for \ac{CFM}, \ac{PFM} and $d'$-\acs{PFM} for generation experiments. The same isometry $\varphi_{\boldsymbol{\theta}}$ of the interpolation experiments is used for the \ac{PFM} and $d'$-\acs{PFM}.}
    \label{tab:generation experiments hyperparameters}
    \resizebox{\linewidth}{!}{
    \begin{tabular}{lcccccc}
        \toprule
        & \multicolumn{3}{c}{\textbf{\acs{ARCH}}} & \multicolumn{3}{c}{\textbf{\acs{I-FABP}}}\\ 
        \textbf{Hyperparameter} & \acs{CFM} & \acs{PFM} & $d'$-\acs{PFM} & \acs{CFM} & \acs{PFM} & $d'$-\acs{PFM} \\ 
        \midrule 
        Epochs & $5000$ & $5000$ & $5000$ & $5000$ & $5000$ & $5000$ \\
        Learning Rate & $0.0005$ & $0.0005$ & $0.0005$ & $0.001$ & $0.001$ & $0.0005$ \\
        Scheduler & \texttt{Cosine} & \texttt{Cosine} & \texttt{Cosine} & \texttt{Cosine} & \texttt{Cosine} & \texttt{Cosine} \\
        Minimum Learning Rate & $5.0 \cdot 10 ^{-6}$ & $5.0 \cdot 10 ^{-6}$ & $5.0 \cdot 10 ^{-6}$ & $1.0 \cdot 10 ^{-5}$ & $1.0 \cdot 10 ^{-5}$ & $5.0 \cdot 10 ^{-6}$\\ 
        Train/Test Split & $0.8/0.2$ & $0.8/0.2$ & $0.8/0.2$ & $0.8/0.2$ & $0.8/0.2$ & $0.8/0.2$\\
        Seed & $0$ & $0$ & $0$ & $0$ & $0$ & $0$ \\
        Number of Layers & $10$ & $10$ & $10$ & $10$ & $10$ & $10$ \\
        Hidden Units & $64$ & $64$ & $16$ & $131\cdot3+1$ & $131\cdot3+1$ & $131\cdot3+1$\\
        Batch Size & $64$ & $64$ & $64$ & $64$ & $64$ & $64$\\
        $n_{\text{simulation steps}}$ & $10$ & $10$ & $10$ & $10$ & $10$ & $10$ \\
        \bottomrule
    \end{tabular}}
\end{table}

\subsection{Designable Latent Manifolds for Novel Protein Engineering}
In \autoref{tab:protein design experiments hyperparameters rae} one can find the settings for training the isometry on the \ac{GRAMPA} dataset for the protein sequence design experiments. Specific hyperparameter worth mentioning is the embedding dimensions, we use an embedding layer from the \texttt{Flax} library to embed the discrete sequences into a continuous space and use a sign-cosine positional embedding, to embed the location in the sequence of the amino acids in the data. 

\begin{table}[h!]
    \centering
    \small 
    \caption{Hyperparameter settings for protein design experiments of the \acp{RAE} on the \ac{GRAMPA} dataset.}
    \label{tab:protein design experiments hyperparameters rae}
    \begin{tabular}{lc}
        \toprule
        \textbf{Hyperparameters} & \textbf{Setting}\\
        \midrule
        Epochs & $1000$ \\ 
        Learning Rate & $0.0001$ \\
        Optimizer & \texttt{Adam} \\
        Train/Test Split & $0.8/0.2$ \\ 
        $n_{steps}$ & $10$ \\ 
        Seed & $0$ \\ 
        Number of Layers & $5$ \\ 
        Embedding dimension & $8$\\
        $\alpha_1$ & $5.0$ \\ 
        $\alpha_2$ & $5.0$ \\ 
        $\alpha_3$ & $5.0$ \\ 
        $\alpha_4$ & $0.05$ \\ 
        $d'$ & $128$ \\ 
        Hidden Units & $512$ \\ 
        Batch Size & $128$ \\ 
        Warmup & $100$ \\ 
        \bottomrule
    \end{tabular}
\end{table}